\documentclass{article}

\usepackage{graphicx}
\usepackage{amssymb, amsmath, amsthm,mathtools}
\usepackage{epstopdf}

\usepackage[marginratio=1:1,height=600pt,width=460pt,tmargin=100pt]{geometry}
\usepackage{layout, color}
\usepackage{enumerate}
\usepackage{authblk} 

\usepackage{array} 
\usepackage[font=small]{caption}

\usepackage{algorithm}
\usepackage{algpseudocode}

\usepackage{setspace}

\usepackage{hyperref}

\usepackage{url}
\usepackage[caption=false]{subfig}
\usepackage{mdwlist}
\usepackage{multirow}
\usepackage{amsthm}
\usepackage{color}

\newcommand{\R}{\mathbb{R}} 
\newcommand{\E}{\mathbb{E}}
\renewcommand{\Pr}{\mathbb{P}}

\newcommand{\MMD}{\rm MMD}
\newcommand{\NTK}{\rm NTK}
\newcommand{\thres}{\rm thres}
\newcommand{\level}{\rm level}
\newcommand{\net}{\rm net}
\renewcommand{\Pr}{\mathbb{P}}

\newcommand{\calN}{\mathcal{N}}

\newcommand{\calD}{\mathcal{D}}

\newcommand{\calX}{\mathcal{X}}

\newtheorem{theorem}{Theorem}[section]
\newtheorem{proposition}[theorem]{Proposition}
\newtheorem{lemma}{Lemma}[section]

\theoremstyle{remark}
\newtheorem{remark}{Remark}[section]

\newtheorem{conjecture*}{Conjecture}

\theoremstyle{plain}

\usepackage[dvipsnames]{xcolor}

\newcommand{\rev}[1]{\textcolor{black}{#1}}

\begin{document}

\title{Neural Tangent Kernel Maximum Mean Discrepancy}

\author{
Xiuyuan Cheng\thanks{Department of Mathematics, Duke University. Email: xiuyuan.cheng@duke.edu}
~~~~~~~~~
Yao Xie\thanks{H. Milton Stewart School of Industrial and Systems Engineering,
Georgia Institute of Technology. Email: yao.xie@isye.gatech.edu.  
}
}

\date{\vspace{-20pt}}

\maketitle

\begin{abstract}
We present a novel neural network Maximum Mean Discrepancy (MMD) statistic by identifying a new connection between neural tangent kernel (NTK) and MMD. This connection enables us to develop a computationally efficient and memory-efficient approach to compute the MMD statistic and perform NTK based two-sample tests towards addressing the long-standing challenge of memory and computational complexity of the MMD statistic, which is essential for online implementation to assimilating new samples.
Theoretically, such a connection allows us to understand the NTK test statistic properties, such as the Type-I error and testing power for performing the two-sample test, by adapting existing theories for kernel MMD. Numerical experiments on synthetic and real-world datasets validate the theory and demonstrate the effectiveness of the proposed NTK-MMD statistic.
\end{abstract}

\section{Introduction}

Maximum Mean Discrepancy (MMD) statistic is a popular method in machine learning and statistics. 
In particular,
kernel MMD \cite{anderson1994two,gretton2012kernel} has been applied to evaluating and training neural network generative models \cite{li2015generative,sutherland2016generative,li2017mmd,arbel2019maximum,kubler2020learning}.
Though a widely used non-parametric test \cite{gretton2012kernel}, kernel MMD
encounters several challenges in practice. The roadblocks for large-scale implementation of kernel MMD involve heavy memory requirement (due to the computation and storage of the Gram matrix, which grows quadratically with the data size) and the choice of a good kernel function \rev{for} high dimensional data. While Gaussian RBF kernel was shown to provide a metric between pairs of probability distributions with infinite data samples, applying isotropic Gaussian kernel to data in applications,
such as image data and discrete events data, may invoke issues in terms of kernel expressiveness \cite{gretton2012optimal,jitkrittum2016interpretable,liu2020learning} and sampling complexity \cite{ramdas2015decreasing}.

A potential path forward in developing more computationally and memory-efficient testing statistics is to leverage deep neural networks' representation and optimization advantage. 
For example,  the idea of training a classification neural network for testing problems has been revisited recently in \cite{lopez2017revisiting,cheng2019classification}, and the connection between classification and two-sample testing 
dates back to earlier works  \cite{friedman2004multivariate, sriperumbudur2009integral,reid2011information}.
However, in applying deep models to testing problems, the test consistency analysis is usually incomplete due to the lack of optimization guarantee of the trained network. For one thing,
assuming perfect training of a deep network to achieve global minimizer is too strong an assumption to fulfill in practice. 

A recent focus of neural network optimization research is the so-called lazy training regime of over-parametrized neural networks \cite{chizat2019lazy}, where the neural network training dynamics exhibit certain linearized property and provable learning guarantee can be obtained \cite{li2018learning,du2018gradient,du2019gradient,allen2019convergence}. In this regime, the training time is sufficiently short, and networks are sufficiently parametrized such that network parameters stay close to the randomized initial values over the training process. In particular, the Neural Tangent Kernel (NTK) theory, as firstly described by \cite{jacot2018neural}, shows that the network optimization can be well approximated by the Reproducing Kernel Hilbert Space (RKHS) formulation.
The NTK theory has been developed for general neural network architectures, including deep 
 fully connected networks \cite{zou2020gradient}, 
convolutional networks \cite{arora2019exact,li2019enhanced}, 
graph neural networks \cite{du2019graph}, and residual networks \cite{tirer2020kernel,huang2020deep,belfer2021spectral}.
The RKHS approach by NTK  has been shown theoretically and empirically to characterize the wide neural network training dynamic in the early stage.

The current work stems from a simple observation that short-time training of a network is approximately equivalent to computing the {\it witness function} of a kernel MMD with NTK at time zero, when the training objective equals the difference between sample averages of the network function on two samples. 
The proposed test statistic, called NTK-MMD,
approximates the classical kernel MMD with NTK, 
and the error at training time $t$ can be bounded to be $O(t)$ under the linearization of the NTK theory. 
The theoretical benefit of translating the network-based statistic into a kernel MMD is that the testing power of the latter can be analyzed based on previous works. 
Algorithm-wise, the network-based test statistic can be computed on the fly: thanks to the form of linear accumulation of the training objective, the training allows small-batch, e.g., batch size 1 and 1 epoch of training (1 pass of the samples), under the NTK approximation.
To calibrate the testing threshold needed to prevent false alarm, we introduce an asymmetric MMD using training-testing split and theoretically prove the testing power where the threshold is estimated from bootstrapping on the test split only and thus avoids retraining network. 

Our main contributions include the following:
(i) We introduce a neural network-based test statistic called NTK-MMD, which can be computed by a short-time training of a neural network, particularly online learning using one-pass of the training samples and batch size one. The NTK approximation error of the MMD statistic is shown to be linear in training time, which leads to the construction of the NTK-MMD test statistic; (ii) We characterize the statistical properties of the NTK-MMD, including the Type-I error and the testing power, which establish the conditions under which the test is powerful; we further introduce a data split scheme such that the test threshold can be estimated without network retraining with provable testing power guarantee; 
(iii) The efficiency of the proposed NTK-MMD test is demonstrated on simulated and real-world datasets.

At the same time, we are aware of the limitations of NTK in explaining deep network optimization, expressiveness power, and so on. We discuss limitations and extensions in the last section. In particular, this paper focuses on demonstrating the power of NTK-MMD statistics for the two-sample test, while the proposed computationally and memory-efficient NTK-MMD statistics can also be used for other applications of MMD statistics other than the two-sample testing problem \cite{gretton2017new}.

\section{Method}\label{sec:method}

\subsection{Preliminary: Kernel MMD}

We start by reviewing a few preliminaries. 
Consider data in $\calX \subset \R^d$, sampled from two unknown distributions with densities $p$ and $q$. 
Given two data sets
\begin{equation}
X = \{ x_i \sim p,  \, {\rm i.i.d.}, \, i =1, \cdots, n_X\},
 \quad
 Y = \{ y_j \sim q,  \, {\rm i.i.d.}, \, j =1, \cdots, n_Y \},
\end{equation}
we would like to test whether or not they follow the same distribution. This is equivalent to perform the following hypothesis test
$H_0: p = q$ versus $H_1: p \neq q$.
The classical kernel MMD considers test functions in the RKHS
of positive semi-definite kernel $K(x,y)$, 
which can be, for instance, the Gaussian RBF kernel. 
The (squared and biased) empirical kernel MMD statistic is given by  \cite{gretton2012kernel}:
\begin{equation}\label{eq:MMD}
{\MMD}^2_K= 
\int_{\calX} \int_{\calX} K(x,y) (\hat{p} - \hat{q})(x) (\hat{p} - \hat{q})(y) dx  dy,
\quad
\hat{p} := \frac{1}{n_{X}} \sum_{i=1}^{n_{X}} \delta_{x_i},
\quad
\hat{q} := \frac{1}{n_{Y}} \sum_{i=1}^{n_{Y}} \delta_{y_i}.
\end{equation}
The null hypothesis is rejected if ${\rm MMD}^2_K > t_{\thres}$,
where $t_{\thres}$ is the user-specified test threshold (usually, chosen to control the false alarm up to certain level). 
The (empirical) {\it witness function} of the MMD statistic,
$\hat{w}(x) = \int_{\calX} K(x,y) (\hat{p} - \hat{q})(y)dy$, indicates where the two densities differ.
The Type-I error of the test is defined as $\Pr[  {\rm MMD}^2_K > t_{\thres} ]$ under $H_0$,  
and the Type-II error as $\Pr[  {\rm MMD}^2_K  \le t_{\thres} ]$ under $H_1$; 
the power is defined as one minus the Type-II error. 
For an alternative distribution $q$ of $p$,
the test errors depend on $q$, the sample sizes $n_X$ and $n_Y$, as well as the kernel function $K(x,y)$. 
Theoretically, the 
test power of kernel MMD has been analyzed in \cite{gretton2012kernel},
investigated for high dimensional Gaussian data in \cite{ramdas2015decreasing},
and for manifold data in \cite{ChengXie2021}.

\subsection{NTK-MMD statistic}

As the proposed NTK-MMD framework can be used on different network architectures, 
we write the neural network mapping abstractly as $f(x; \theta)$, which maps from input $x \in \calX$ to $\R$,
and $\theta$ is the network parameters. 
Use $X \cup Y$ as the training dataset, and let $\hat{p}$ and $\hat{q}$ be as in \eqref{eq:MMD}, we choose a particular  training objective function \rev{as}
\begin{equation}\label{eq:def-Loss}
\hat{L}(\theta) 
= -\int_{\calX} f( x; \theta) (\hat{p} -  \hat{q})(x) dx
= - \frac{1}{n_X}\sum_{i=1}^{n_X} f(x_i; \theta) +  \frac{1}{n_Y}\sum_{i=1}^{n_Y} f(y_i; \theta)
\end{equation}
 The choice of this objective function is critical in establishing the connection between NTK and MMD. Optimizing this objective will lead to divergence of the network function if we train for a long time. However,
if the training is only for a short time, 
the network function remains finite. Here we mainly focus on short-time training of the network, and particularly the online training setting where the number of epochs is 1, that is, only 1-pass of the training samples is used. We will also show that the method allows using minimal batch size for online learning without affecting the NTK approximation of the MMD statistic, c.f. Remark \ref{rk:SGD}. 

Following the  convention in NTK literature, 
below we formulate in terms of continuous-time Gradient Descent (GD) dynamic of the network training.
The extension to discrete-time Stochastic Gradient Descent (SGD) holds with a small learning rate (Remark \ref{rk:SGD}).
The network parameter $\theta(t)$ evolves according to $\dot{\theta}(t) = - \partial \hat{L}/\partial \theta$,
and we define $u(x,t) := f(x, \theta(t))$,
which is the network mapping function at time $t$. 
Suppose the network is trained 
for a short time $ t > 0$, 
we define 
\begin{equation}\label{eq:def-g}
 \hat{g}(x): =  \frac{1}{t} (u(x,t) - u(x,0)),
\end{equation}
and the test statistic, which depends on time $t$, is 
\begin{equation}\label{eq:def-hatT-net}
\hat{T}_{\rm net}(t) := \int_{\calX} \hat{g}(x)  (\hat{p} -  \hat{q}) (x) dx
\rev{ = \frac{1}{t} \left( \hat{L}(\theta(t)) -  \hat{L}(\theta(0)) \right)}.
\end{equation}
The function $\hat{g}$ is the difference of the network mapping after a short-time training from the initial one,
and we call it the {\it witness function} of network NTK-MMD statistic.
\rev{
As revealed by \eqref{eq:def-hatT-net},
(without calibrating the test threshold)
the test statistics $\hat{T}_{\rm net}$ is nothing but the decrease in the training objective, 
and comes as a by-product of network training at no additional computational cost.}
We show in next subsection that at small $t$,
the statistic $\hat{T}_{\net}(t)$ \rev{provably} approximates the classical MMD statistic with the NTK,
i.e.
$\hat{T}_{\NTK} =  {\MMD}^2_{K} (X, Y) 
$
where $K$ is the NTK at time $t=0$ as in \eqref{eq:def-K0}.
\rev{Algorithmically,}
we will perform two-sample test using $\hat{T}_{\net}(t)$ by comparing with a threshold $t_{\rm thres}$.

\subsection{NTK approximation of MMD statistic}\label{subsec:ntk-approx-error}

In the continuous-time training dynamic of the network, 
\rev{we consider the NTK \cite{jacot2018neural} kernel function defined for $t > 0$ as}
\begin{equation}\label{eq:def-hatKt}
\hat{K}_t(x,x') := \langle \nabla_\theta f( x; \theta(t)), \nabla_\theta f( x'; \theta(t)) \rangle.
\end{equation}
The following lemma follows directly by construction, and the proof is in \rev{Appendix \ref{appsub:proofs-sec2}}.
\begin{lemma}\label{lemma:u(x,t)}
The network function $u(x,t)$ satisfies that for $t> 0$,
\begin{equation}\label{eq:u-evolve}
u(x,t)-u(x,0) = \int_0^t  \int_\calX  \hat{K}_s(x,x')   (\hat{p} - \hat{q})(x') dx' ds.
\end{equation}
\end{lemma}

It has been shown \rev{(in \cite{arora2019exact,arora2019fine,chizat2019lazy}, among others)} that for the short-time training (lazy training regime), 
the kernel \eqref{eq:def-hatKt} can be well-approximated by the kernel at time $t=0$, namely
\begin{equation}\label{eq:def-K0}
K_{0}(x,x') := \langle \nabla_\theta f( x; \theta(0)), \nabla_\theta f( x'; \theta(0)) \rangle,
\end{equation}
which is only determined by the network weight initialization $\theta(0)$. 
Assuming $K_0 \approx \hat{K}_t$ in Lemma \ref{lemma:u(x,t)}, 
the proposed test statistic $\hat{T}_{\rm net}(t)$ as in \eqref{eq:def-hatT-net} 
can be viewed as 
\begin{equation}\label{eq:def-hatT}
\hat{T}_{\rm net} (t) \approx \hat{T}_{\NTK} := 
\int_{\calX} 
 \int_{\calX} K_{0}(x,x')
 (\hat{p} -  \hat{q} ) (x) 
 (\hat{p} -  \hat{q} )(x')dx dx',
\end{equation}
which is the kernel MMD statistic with NTK. 
\rev{(See  Remark \ref{rk:unbiased-mmd} for a discussion on biased/unbiased MMD estimator.)}
\rev{In below, we show in Proposition \ref{prop:ntk-O(t)}}
that the approximation $\hat{T}_{\net} \approx \hat{T}_{\NTK}$ has $O(t)$ error,
and we experimentally verify the similarity of the two statistics in Subsection \ref{subsec:exp-ntk}.
Throughout the paper, we compute $\hat{T}_{\rm net}$ by neural network training, 
and we call $\hat{T}_{\rm NTK}$ the {\it exact} NTK-MMD which is for theoretical analysis.
The theoretical benefit of translating $\hat{T}_{\net}$ into $\hat{T}_{\NTK}$ lies in that 
testing power analysis of $\hat{T}_{\NTK}$ follows existing methods
which is detailed in Section \ref{sec:theory}.

Suppose neural network parameter $\theta$ is in $\R^{M}$ and $\theta \in \Theta$,
where $\Theta$ is a domain in $\R^M$ which contains the Euclidean ball $B(\theta(0), r_0)$,
where we assume $r_0$ is an $O(1)$ constant. 
For vector valued  function $g: (\calX, \Theta) \to \R^d$ and $U \subset \Theta$, we denote the infinity norm as
$\|g\|_{\calX, U} := 
\sup_{x \in \calX, \theta \in U} \| g (x,\theta) \|$.
\rev{When $g$ maps to a matrix, the notation denotes (the infinity norm over $(\calX \times U)$ of) the operator norm.}
The test statistic approximation error in Proposition \ref{prop:ntk-O(t)} directly follows the following lemma concerning the uniform approximation of the kernels. 
All proofs in \rev{Appendix \ref{appsub:proofs-sec2}}.

\begin{lemma}[NTK kernel approximation]
\label{lemma:ntk-approx}
Suppose $f$ is $C^2$ on $(\calX, \Theta)$ 
and $\| \nabla_\theta f  \|_{\calX, \Theta} \le L_f$ for some positive constant $L_f$.
Then for any \rev{$0 < r <r_0 $}, when $ 0 < t < t_{f,r}: = r/(2 L_f)$,\\
(1) $\theta(t)$ stays inside the Euclidean ball $B_r : = B(\theta(0), r)$.\\
(2) Define $C_{f,r}:= 4 \| {\rm D}_\theta^2 f\|_{\calX, B_r} \| \nabla_\theta f\|_{\calX, B_r}^2$,  we have that
\begin{equation}
\sup_{x, x' \in \calX} | \hat{K}_t(x,x') - K_{0}(x,x')  | \le C_{f,r} t.
\end{equation}
\end{lemma}

\begin{remark}[Boundedness of $\| \nabla_\theta f  \|_{\calX, \Theta} $]
When $p$ are unbounded density (gaussian),
and activation function $f$ is relu or softplus,
the uniform boundednesss of $\| \nabla_\theta f  \|_{\calX, \Theta}$ may fail.
However, for sub-exponential densities, apply standard truncation argument,
and when we restrict to compactly supported distributions.
In practice, we standardize the data to be on a compact domain in $\R^d$.
\end{remark}

\begin{proposition}[Test statistic approximation]
\label{prop:ntk-O(t)}
The condition on $f(x,\theta)$ is the same as in Lemma \ref{lemma:ntk-approx}, 
and  for \rev{$0 < r <r_0 $}, the constants $t_{f,r}$ and $C_{f,r}$ are as therein.
Then, when $ 0 < t < t_{f,r}$, we have that
\[
| \hat{T}_{\rm net}(t) - \hat{T}_{\NTK }| \le 2 C_{f,r} t. 
\]
\end{proposition}

\begin{remark}[SGD and online training]\label{rk:SGD}
The above error bound analysis based on Taylor expansion can extend to discrete-time GD dynamic
by showing that the time discretization introduces higher-order error when $t$ is small. 
In the SGD setting, e.g., the online learning of 1 epoch, batch size one, and learning rate $\alpha$, 
the network parameters are updated after scanning each training sample on the fly.
Let $\theta_k$ be the network after scanning $k$ many samples,
we show in Appendix \ref{subsec:sgd-bound} that 
the difference $\|\theta_k - \theta_0 \|$ can be bounded by $O(\alpha k /n)$,
and the trained network witness function  after 1 epoch
approximates the witness function with the zero-time NTK kernel up to an $O(\alpha)$ error.
The learning rate $\alpha$ has the role of training time $t$. 
The fact that batch size will not affect the NTK approximation of the network training is a result of that the loss 
\eqref{eq:def-Loss} is a linear accumulation over samples,
which  may not hold for other loss types. 
The compatibility with online learning and training with very small batch size of NTK-MMD statistic 
makes it convenient for deep network training, especially under memory constraints.
\end{remark}

\subsection{Computational and memory efficiency}

The update of network parameters in NTK-MMD training can be viewed as an implicit computation
of  the inner-product between high dimensional kernel feature maps $\langle \nabla_\theta f( x; \theta(t)), \nabla_\theta f( x'; \theta(t)) \rangle$ 
(by chain rule, c.f. \eqref{eq:partial-t-f-eqn1.1} \eqref{eq:partial-t-f-eqn1} in Appendix \ref{appsub:proofs-sec2}).
The network witness function $\hat{g}$ defined in \eqref{eq:def-g} is parametrized  and stored in trained network parameters. 
This allows the (approximate) evaluation of kernel on a test sample $x'$ without computing the gradient $\nabla_\theta f( x'; \theta)$ explicitly.
It also means that the NTK network witness function can be evaluated on any new $x'$ without revisiting the training set.
In contrast, traditional kernel MMD computes kernel witness function
(defined as $\int K(x,y) (\hat{p}-\hat{q})(y)dy$ \cite{gretton2012kernel})
 on a new point $x'$ by pairwise computation between $x'$ and samples in datasets $X$ and $Y$.

NTK-MMD can be computed via batch-size-one training over one-pass of the training set 
(c.f. Remark \ref{rk:SGD} and experimentally verified in Table \ref{tab:exp-more-sgd}).
The gradient field evaluation (back propagation) is only conducted on the training set but not the testing test, 
and the bootstrap calibration of the test threshold can be computed from test set only (c.f. Section \ref{subsec:hatT-a}). 
Thus, by using small learning rate (allowed by floating point precision, c.f. Remark \ref{rk:effective-lr}), 
one can incorporate large number of  training samples via more training iterations without worsening the approximation error to exact NTK-MMD,
which will improve testing power. 
This ``separation'' of training and testing, in memory and computation,
of NTK-MMD allows scalable online learning as well as efficient deployment of the network function on potentially large test sets.

\section{Theoretical properties of NTK-MMD}\label{sec:theory}

In this section, we prove the testing power (at a controlled level) 
of the NTK-MMD statistic $\hat{T}_{\rm NTK}$ as in \eqref{eq:def-hatT}
with large enough finite samples.
We also introduce an asymmetric version of the MMD statistic using training-testing dataset splitting,
which enables the bootstrap estimation of the threshold of the test $t_{\thres}$ without retraining of the neural network.

\subsection{NTK-MMD without data splitting}\label{subsec:hatT}

We write the NTK kernel $K_0(x,x')$ as $K(x,x')$ omitting the $_{NT}$ subscript,
and assume that $K(x,x')$ is uniformly bounded, that is,
$\sup_{x \in \calX } K(x,x) \le B < \infty$ for some positive constant $B$.  
Without loss of generality, 
we assume that $B=1$
(because a global constant normalization of the kernel does not change the testing). 
By that the kernel is PSD, we thus have that
\begin{equation}\label{eq:Kernel-uniform-bound}
\sup_{x', \, x \in \calX }| K(x, x') | \le 1. 
\end{equation}
We omit the $_{\rm NTK}$ subscript and denote the MMD statistic \eqref{eq:def-hatT}  as $\hat{T}$.
The corresponding population statistic is the squared MMD of kernel $K(x,y)$
\begin{equation}\label{eq:def-delta-K}
\delta_K: = 
{\rm MMD}_K^2 (p,q) = \int_\calX \int_\calX K(x,y)(p-q)(x)(p-q)(y) dxdy.
\end{equation}
By the uniform boundedness  \eqref{eq:Kernel-uniform-bound},  
the kernel $K(x,x')$ is in $L^2(\calX \times \calX, (p+q)(x)(p+q)(x')dx dx')$.
We define the squared  integrals of the kernel
\begin{equation}\label{eq:def-nupp-and-all}
\begin{split}
\nu_{pp} &:= \E_{x \sim p, y \sim p }K(x,y)^2, 
\quad
\nu_{pq} :=  \E_{x \sim p, y \sim q } K(x,y)^2,  
\quad
\nu_{qq}  :=  \E_{x \sim q, y \sim q }  K(x,y)^2.
\end{split}
\end{equation}
In addition, we assume that as $n:= n_X + n_Y$ increases, 
$n_X/n $ stay bounded and approaches $\rho_{X} \in (0,1)$.
Equivalently, there is some constant $ 0 < c < 1$ such that for large enough $n$,  
\begin{equation}\label{eq:def-cn-balance}
c n +1   \le n_X, \, n_Y \le n, \quad i=1,2.
\end{equation}
Without loss of generality,  we assume that \eqref{eq:def-cn-balance} always holds for the $n$ considered.

\begin{theorem}[Test power of $\hat{T}_{\rm NTK}$]
\label{thm:power}
Suppose \eqref{eq:Kernel-uniform-bound} and  \eqref{eq:def-cn-balance} hold, and

(i) Under $H_1$, $p \neq q$, 
the squared population kernel MMD $\delta_K $ as in \eqref{eq:def-delta-K} is strictly positive, 

(ii)  The three integrals as in \eqref{eq:def-nupp-and-all}, 
$\nu_{pp}, \nu_{pq}, \nu_{qq}$,
are all bounded by a constant $\nu \le 1 $.

Define 
$
\lambda_1 := \sqrt{8 \log (4/\alpha_{\rm level})}$,
and let the threshold for the test be 
$
t_{\thres} =  4/(cn) + 4 \lambda_1 \sqrt{\nu/cn }$.
Then, if for some $\lambda_2 > 0$, $n$ is large enough such that 
\begin{equation}\label{eq:cond-n}
n > \frac{1}{c} \max  
\left\{ 
\frac{1}{ 9 \nu}   \max\{ \lambda_1 , \lambda_2 \}^2, \, 
\frac{8}{\delta_K}, \,
\frac{\nu}{ \delta_K^2} \left(  8(\lambda_1 + \lambda_2) \right)^2 \right \},
\end{equation}
then under $H_0$,
$\Pr [\hat{T} > t_{\thres}] \le \alpha_{\rm level}$;
and 
under $H_1$,
$\Pr [\hat{T} \le t_{\thres}] \le 3 e^{-\lambda_2^2/8} $.
\end{theorem}

The proof uses the U-statistic concentration analysis, and is left to Appendix \ref{appsec:proof-sec-theory}.
As revealed by the proof, 
the diagonal entries in the kernel matrix contribute to the $O(1/n)$ term,
and thus switching from the biased  estimator of MMD  \eqref{eq:def-hatT} to the unbiased estimator gives similar theoretical results.

\begin{remark}[Choice of $t_{\thres}$]\label{rk:choice-t-thres}
The choice of $t_{\thres}$ in the above theorem is a theoretical one and may not be optimal,
\rev{due to the use of concentration inequality and the relaxation of the bounds by using constants $\nu$ and $c$}. 
By definition, the optimal value of $t_{\thres}$  is the (1 - $\alpha_{\rm level}$)-quantile of the distribution of $\hat{T}$ under $H_0$. 
\rev{The asymptotic choice may be obtained analytically according to the limiting distribution of the MMD statistic, c.f. Remark \ref{rk:asymptotic-t-thres}}.
The threshold $t_{\thres}$ is also computed by
 a bootstrap strategy in practice \cite{arcones1992bootstrap} (called ``full-bootstrap'' in next subsection).
The bootstrap approach permutes the labels in data sets $X$ and $Y$,
 and since in $\hat{T}_{\rm net}$ the witness function $\hat{g}(x)$ is computed by neural network training,
 this will incur retraining of the network.
A solution to avoid retraining by adopting a test set for bootstrap estimation of $t_{\thres}$ is introduced in next subsection.  
\end{remark}

\subsection{Threshold calibration by data splitting}\label{subsec:hatT-a}

As shown in Theorem \ref{thm:power} and Remark \ref{rk:choice-t-thres},
in the theoretical characterization of test power (at a required test level)
the test threshold plays a critical role.
In practice, we need a more precise threshold to exactly control the false alarm under the null hypothesis.
 In this section, we discuss how to set the threshold in two settings: fixed-sample and pilot-data. Nevertheless, we would like to mention that there exist applications where the threshold is not needed, and the symmetric MMD $\hat{T}_{\rm NTK}$ can be used as a measurement of \rev{distribution} divergence.

{\bf Fixed-sample setting.} We first consider the setting where we have a fixed number of samples from $p$ and $q$. To obtain a precise threshold to control the false alarm, we need to split data into two non-overlapping parts: one part for training neural networks (compute the witness function) and one part data for bootstrapping and calibrating the threshold. We want to highlight that here we develop a scheme for threshold calibration such that no re-training of the witness function is necessary.

We randomly split the datasets $X$ and $Y$ into training and testing sets,
$X = X_{(1)} \cup X_{(2)}$ and 
$Y = Y_{(1)} \cup Y_{(2)}$, 
and compute an asymmetric version of kernel MMD (the subscript $_a$ is for ``asymmetric'') 
\begin{equation}\label{eq:hatT-a}
\hat{T}_{a} := \int_\calX \int_\calX K(x, x')(\hat{p}_{(1)}- \hat{q}_{(1)})(x') (\hat{p}_{(2)}- \hat{q}_{(2)})(x) dx dx',
\end{equation}
where $\hat{p}_{(i)}$ and $\hat{q}_{(i)}$ are the empirical measures of datasets $X_{(i)}$ and $Y_{(i)}$ respectively,
$i=1,2$. Define $n_{X, (i)} = |X_{(i)}|$ and $n_{Y, (i)} = |Y_{(i)}|$, $i=1,2$. 
Similarly as in Section \ref{sec:method}, 
 the MMD statistic \eqref{eq:hatT-a}  with $K(x,y) = K_0(x,y)$, the zero-time NTK,
 can be approximated by
 \begin{equation}\label{eq:hatT-a-net}
 \hat{T}_{a, {\rm net}} (t)= \int_\calX \hat{g}_{(1) }(x) (\hat{p}_{(2)}- \hat{q}_{(2)})(x) dx,
 \quad \hat{g}_{(1) }(x) = \frac{1}{t} ( \hat{u}(x,t)-\hat{u}(x,0) ),
 \end{equation}
 for a small time $t$, 
where $\hat{u}(x,t)$ is the network function trained by 
minimizing 
$\hat{L}(\theta) : = -\int_{\calX} f( x; \theta) (\hat{p}_{(1)} -  \hat{q}_{(1)})(x) dx$
on the training set 
 $\calD_{tr}  = \{ X^{(1)}, Y^{(1)} \}$ with binary labels $\{1, 2 \}$.
 Same as in Lemma \ref{lemma:ntk-approx} Proposition \ref{prop:ntk-O(t)},
 the difference $|\hat{T}_{a} -  \hat{T}_{a, {\rm net}} (t) |$ can be bounded to be $O(t)$.
We  theoretically analyze the testing power of $\hat{T}_a$ where $K(x,y) = K_0(x,y)$ in below. 
 
The benefit of splitting the test set lies in that 
 once the witness function $\hat{g}_{(1)}(x)$ is trained from  $\calD_{tr}$, 
 one can do a {\it test-only bootstrap} which is to compute
 \begin{equation}\label{eq:def-T-null-split}
 \hat{T}_{a, \rm null} = \int_\calX \hat{g}_{(1) }(x) (\hat{p}_{(2)}' - \hat{q}_{(2)}')(x) dx,
 \end{equation}
 where $\hat{p}_{(2)}'$ and  $\hat{q}_{(2)}'$ are empirical measure of samples in 
  $\calD_{te}  = \{ X^{(2)}, Y^{(2)} \}$ by randomly permute the $n_{X,(2)} + n_{Y,(2)}$ many  binary class labels. 
\rev{
Since permuting test labels does not affect $\hat{g}_{(1)}(x)$,
the test-only bootstrap does not require retraining of the network.
}
\rev{Alternatively,}
one can permute the binary class labels in both $\calD_{tr}$ and $\calD_{te}$,
and will require to retain the neural network  to obtain the new witness function $\hat{g}_{(1)}$ given the new class labels of $\calD_{tr}$.
We call such a bootstrap the {\it full-bootstrap}.
The full-bootstrap can be applied to the symmetric MMD statistic without test set splitting as well, \rev{namely the setting of Theorem \ref{thm:power},
to obtain an estimate of optimal $t_{\thres}$.}

We give two theoretical results on the testing power guarantee of the asymmetric NTK MMD statistic \eqref{eq:hatT-a}:
For {\it test-only bootstrap}, Theorem \ref{thm:power-asym-testonly-boot}  proves testing power 
by restricting to good events over the randomness of $\calD_{tr}$;
For {\it full bootstrap}, the guarantee is provided in Theorem \ref{thm:power-asym}, which is the counterpart of Theorem \ref{thm:power}.
All proofs are in \rev{Appendix \ref{appsec:proof-sec-theory}}.

We assume the balance-ness of the two samples as well as the training and testing splitting,
that is, 
$n_{X, (1)} / n_X \to \rho_{X, (1)}$, 
$n_{Y, (1)} / n_Y \to \rho_{Y,(1)}$
$n_X/n \to \rho_{X}$,
and the three constants are all in $(0,1)$.
With $n = n_X + n_Y$,
we assume for constant $0< c  <1$.
\begin{equation}\label{eq:def-cn-balance-split}
c_a n  \le n_{X,(i)}, \, n_{Y,(i)} \le n, \quad i=1,2.
\end{equation}
We denote by $\Pr_{(1)}$ the randomness over $\calD_{tr}$, 
and $\Pr_{(2)}$ that over $\calD_{te}$.

\begin{theorem}[Test power of $\hat{T}_a$, test-only bootstrap]
\label{thm:power-asym-testonly-boot}
Suppose that
\eqref{eq:Kernel-uniform-bound}, 
\eqref{eq:def-cn-balance-split} and the conditions (i) and (ii) in Theorem \ref{thm:power} hold,
and $0< \gamma <1$ is a small number.
Define $\lambda_{(2),1} := \sqrt{ 4 \log (4/\alpha_{\rm level})}$,
$\lambda_{(1)} := \sqrt{ 4 \log (8/\gamma ) }$,
and set the threshold as $t_{\thres} = 4 ( \sqrt{1.1} \lambda_{(2),1} + \lambda_{(1)} ) \sqrt{\nu/(c_an)}$.
If  $n$ is large enough such that
$n > (\frac{\lambda_{(1)}}{0.1 \nu})^2/(8 c_a)$,
and for some $\lambda_{(2),2} > 0$, 
\begin{equation}\label{eq:cond-n-testonly-boot}
n > \frac{1}{c_a} \max  
\left\{ 
\frac{1}{ 9 \nu}   \max\{ 
\lambda_{(1)}, \lambda_{(2),1} , \lambda_{(2),2} \}^2, \, 
\frac{ 16 \nu}{ \delta_K^2} \left(   2\lambda_{(1)} + \sqrt{1.1}(\lambda_{(2),1}+\lambda_{(2),2}) \right)^2 \right \},
\end{equation}
then, under both $H_0$ and $H_1$
there is a good event over the randomness of $\calD_{tr}$ which happens w.p.$\ge 1-\gamma$,
under which, conditioning on $\calD_{tr}$,
$\Pr_{(2)} [\hat{T}_a > t_{\thres}] \le \alpha_{\level}$  under $H_0$,
and
$\Pr_{(2)} [\hat{T} \le t_{\thres}] \le 4 e^{-\lambda_{(2),2}^2/4} $ under $H_1$.
\end{theorem}

\begin{remark}[\rev{Sampling complexity}]
Compared to the full-bootstrap result Theorem \ref{thm:power-asym},
the additional requirement on $n$ is that $c_a n$ needs to be greater than $(\lambda_{(1)}/\nu)^2$ up to absolute constant, and thus when $\nu/\delta_K^2 \gg \nu^{-2}$,  the $(\nu/\delta_K^2)$-term still dominates the needed lower bound of $n$,
same as in  Theorems \ref{thm:power}  and  \ref{thm:power-asym}. 
\rev{
(Here we treat $\lambda_{(1)}$, $\lambda_{(2),1}$ and $\lambda_{(2),2}$ as $O(1)$ constants.
Because the constant $\gamma$ controls the good event probability over the randomness of $\calD_{tr}$,
thus if $\gamma$ can be chosen to be of the same order as $\alpha_{\level}$, then $\lambda_{(1)}$ has the same order as $\lambda_{(2),1}$.)
}
The result shows that with test split and test-only bootstrap (avoiding re-training), 
the test power has the same \rev{order} of needed sampling complexity, \rev{$ n \gtrsim \nu/\delta_K^2$,}
as full bootstrap, with high probability and for large enough $n$. 
\end{remark}

\begin{theorem}[Test power of $\hat{T}_a$, full bootstrap]
\label{thm:power-asym}
Suppose that
\eqref{eq:Kernel-uniform-bound}, 
\eqref{eq:def-cn-balance-split} and the conditions (i) and (ii) in Theorem \ref{thm:power} hold.
Define $ \lambda_1 := \sqrt{8 \log (4/\alpha_{\level})}$,
and let the threshold for the test be 
$t_{\thres} =   4 \lambda_1 \sqrt{ \frac{\nu}{c_a n} }$.
Then, if for some $\lambda_2 > 0$, $n$ is large enough such that 
\begin{equation}\label{eq:cond-n-split}
n > \frac{1}{c_a} \max  
\left\{ 
\frac{1}{ 9 \nu}   \max\{ \lambda_1 , \lambda_2 \}^2, \, 
\frac{\nu}{ \delta_K^2} \left(  4(\lambda_1 + \lambda_2) \right)^2 \right \},
\end{equation}
then under $H_0$,
$\Pr [\hat{T} > t_{\thres}] \le \alpha_{\rm level}$;
and 
under $H_1$,
$\Pr [\hat{T} \le t_{\thres}] \le 4 e^{-\lambda_2^2/8} $.
\end{theorem}

{\bf Pilot data setting.} This section considers the setting where we may have many samples for one distribution, e.g., the $p$. For instance, in change-point detection, where we are interested in detecting a shift in the underlying data distribution, there can be a large pool of pilot data before the change happens, collected historically and representing the normal status. We may have fewer data samples for the distribution $q$. For such a case, we can use data from the reference pool represent distribution $p$ to train the model and calibrate the threshold, e.g., using bootstrap. Since such ``training'' is done offline, we can afford the higher computational cost associated with training the model multiple times.
In short, our strategy is to pre-compute the detector (re-train multiple times) and then use boostrap to obtain the threshold $t_{\thres}$ for detector: (i) compute the symmetric MMD
$\hat{T}_{\rm NTK}$ on $\{ \hat{p}, \hat{q} \}$, where $\hat{q}$ is the new coming test samples (e.g. in change-point detection),
and $\hat{p}$ is from the pool. (ii) pre-compute the symmetric MMD
$\hat{T}_{\rm null}$ on $\{ \hat{p}_2, \hat{p}_2' \}$ from the pool of samples,
with retrain, and obtain the ``true'' threshold for $\hat{T}_{\rm NTK}$. 
Re-training of the network is expensive, but this is pre-computation and not counted in the online computation.

\section{Numerical experiments}

The section presents several experiments to examine the proposed method and validate the theory.
\footnote{Code available at 
\url{https://github.com/xycheng/NTK-MMD/}.}

\subsection{Gaussian mean and covariance shifts}\label{subsec:exp-gaussian}

{\bf Set-up}. Consider Gaussian mean shift and covariance shift in $\R^{100}$, 
where $n_X = n_Y = 200$; $p$ is  the distribution of $\calN(0, I_d)$, $d=100$: (i) Mean-shift: $q$ is the distribution of $\calN(\mu, I_d)$, where $\| \mu \|_2 = \delta$ which varies from 0 to 0.8 and (ii) Covariance-shift: $q $ is the distribution of $\calN(0, I_d + \rho E )$, where $E$ is an $d$-by-$d$ all-ones matrix,
and $\rho$ changes from  0 to 0.16.
We split training and test sets into halves,
and compute the asymmetric network approximated NTK MMD statistic $\hat{T}_{a,\rm net}$ \eqref{eq:hatT-a-net},
and estimate the test threshold by the quantile of \eqref{eq:def-T-null-split};
$H_0$ is rejected if $\hat{T}_{a, \rm net} > t_{\rm thres}$. 
We use a 2-layer network (1 hidden layer) with soft-plus activation. 
The online training is of 1 epoch (1 pass over the training set) with batch-size = 1. 
The bootstrap estimate of test threshold uses $n_{\rm boot} = 400$ permutations. 
The testing power is approximated by $n_{\rm run}=500$ Monte Carlo replicas,
and we compare with the benchmarks by (i) Hotelling's T-test,
and (ii) Gaussian kernel MMD test (median distance bandwidth) \cite{gretton2012kernel}.
The median distance bandwidth is a reasonable choice for detecting high dimensional Gaussian mean shift \cite{ramdas2015decreasing}.
Both Gaussian kernel MMD and Hotelling's Test have access to all the samples $\calD_{tr} \cup \calD_{te}$.
\rev{More experimental details are in Appendix \ref{appsub:exp-gaussian-more}.}

\begin{figure}[t]
\begin{center}
\includegraphics[trim =  0 0 0 0, clip, height=.21\linewidth]{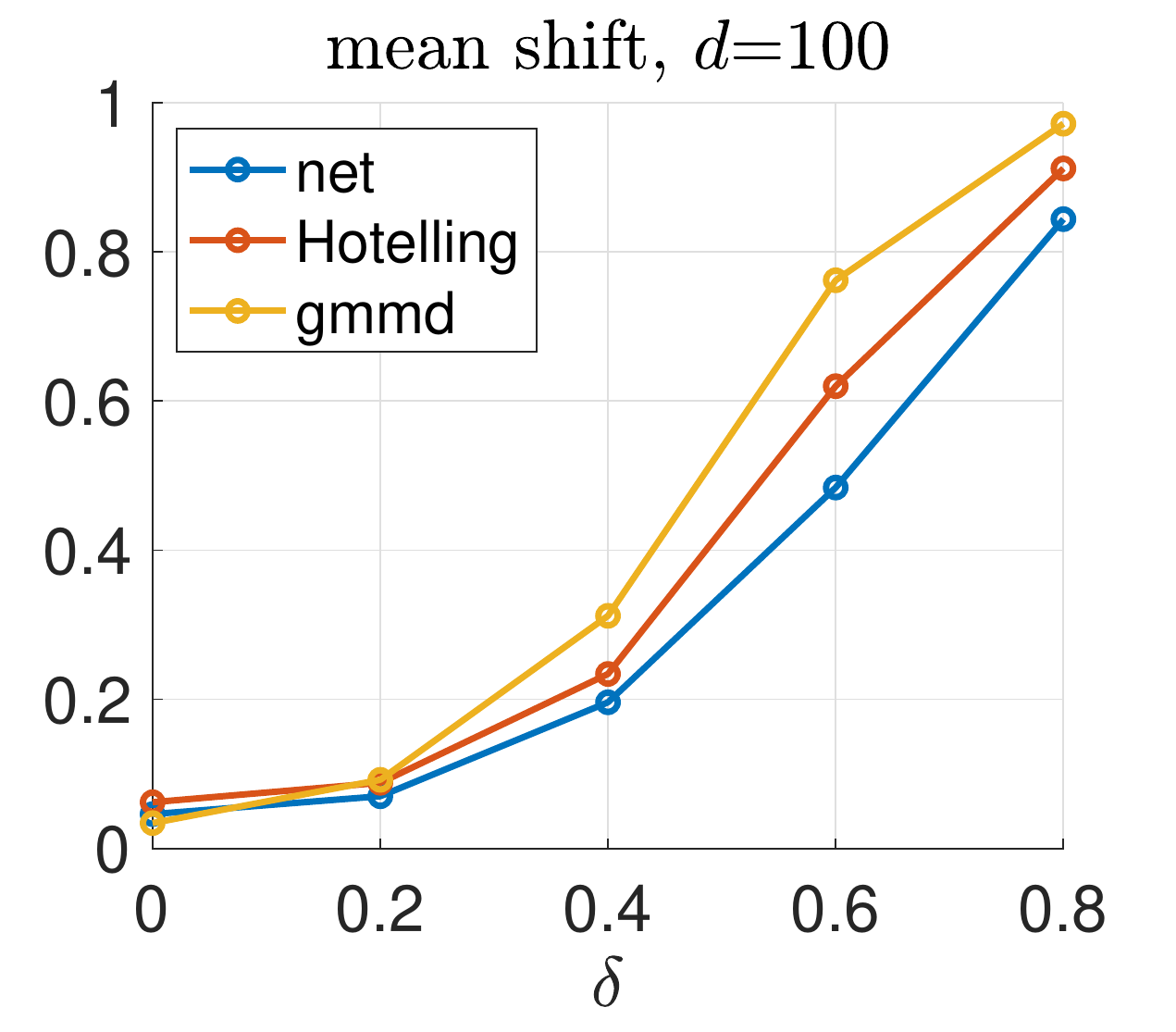} 
\includegraphics[trim =  0 0 0 0, clip, height=.21\linewidth]{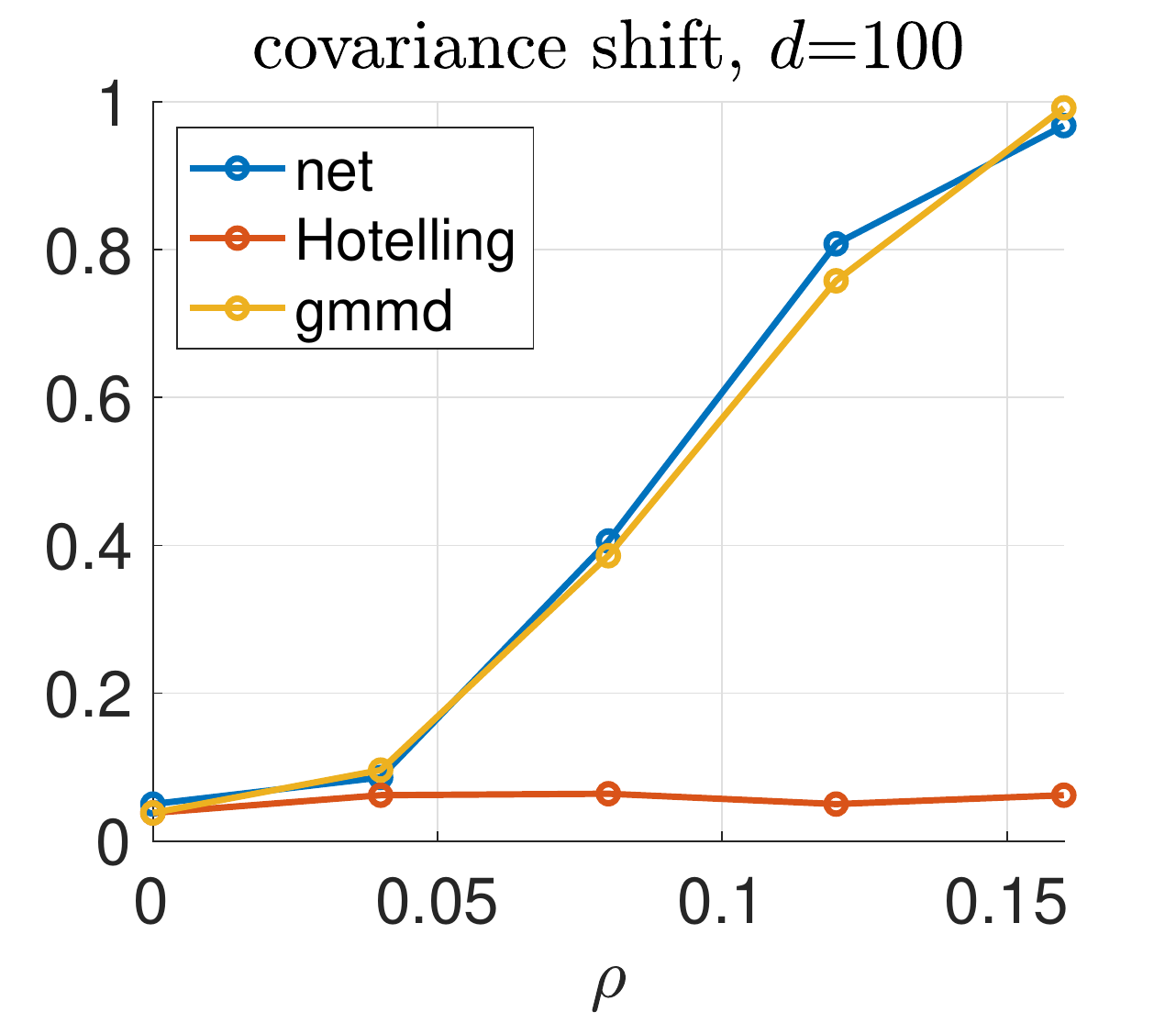} 
\includegraphics[trim =  0 0 0 0, clip, height=.209\linewidth]{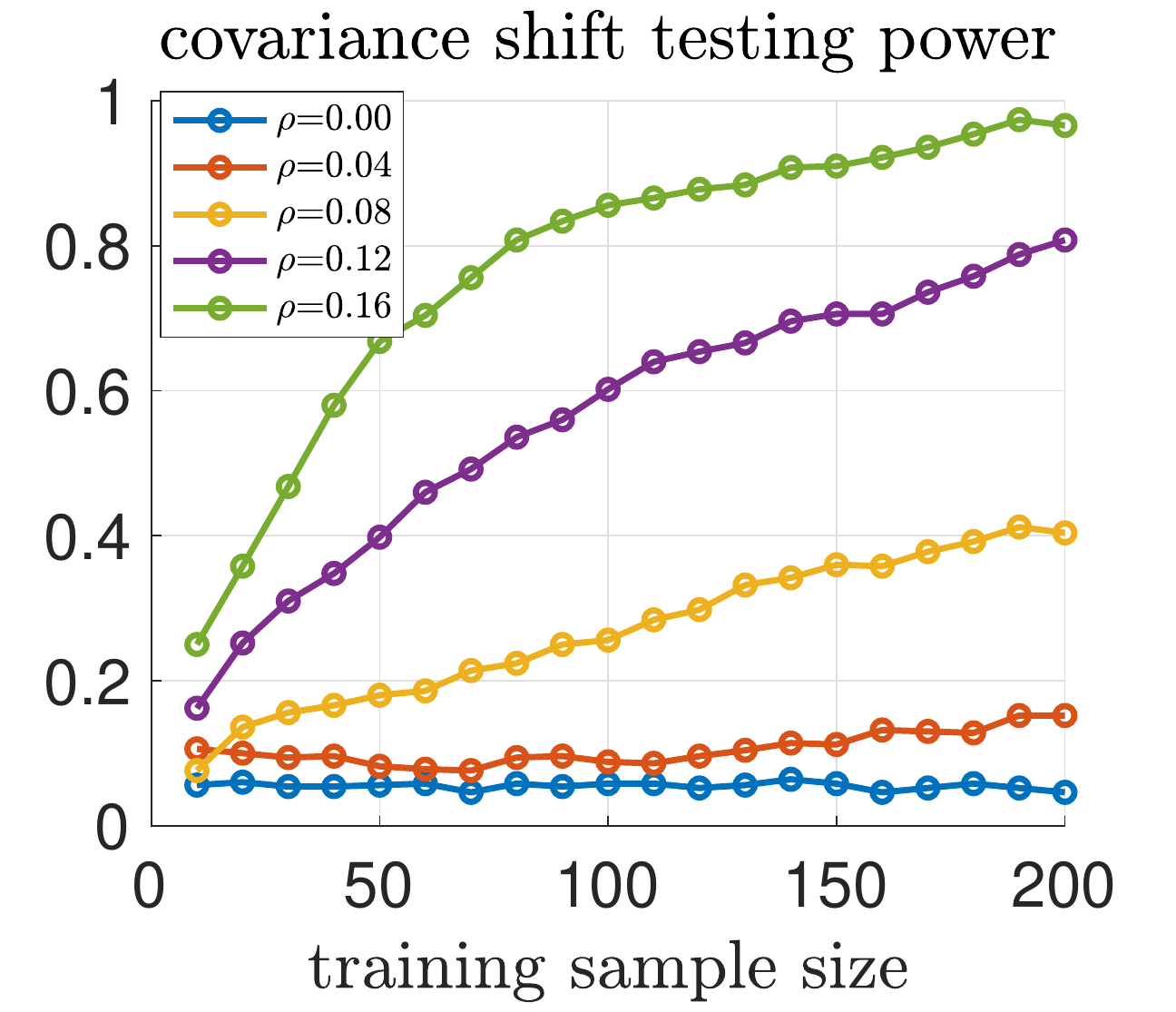} 
\includegraphics[trim =  0 0 0 0, clip, height=.21\linewidth]{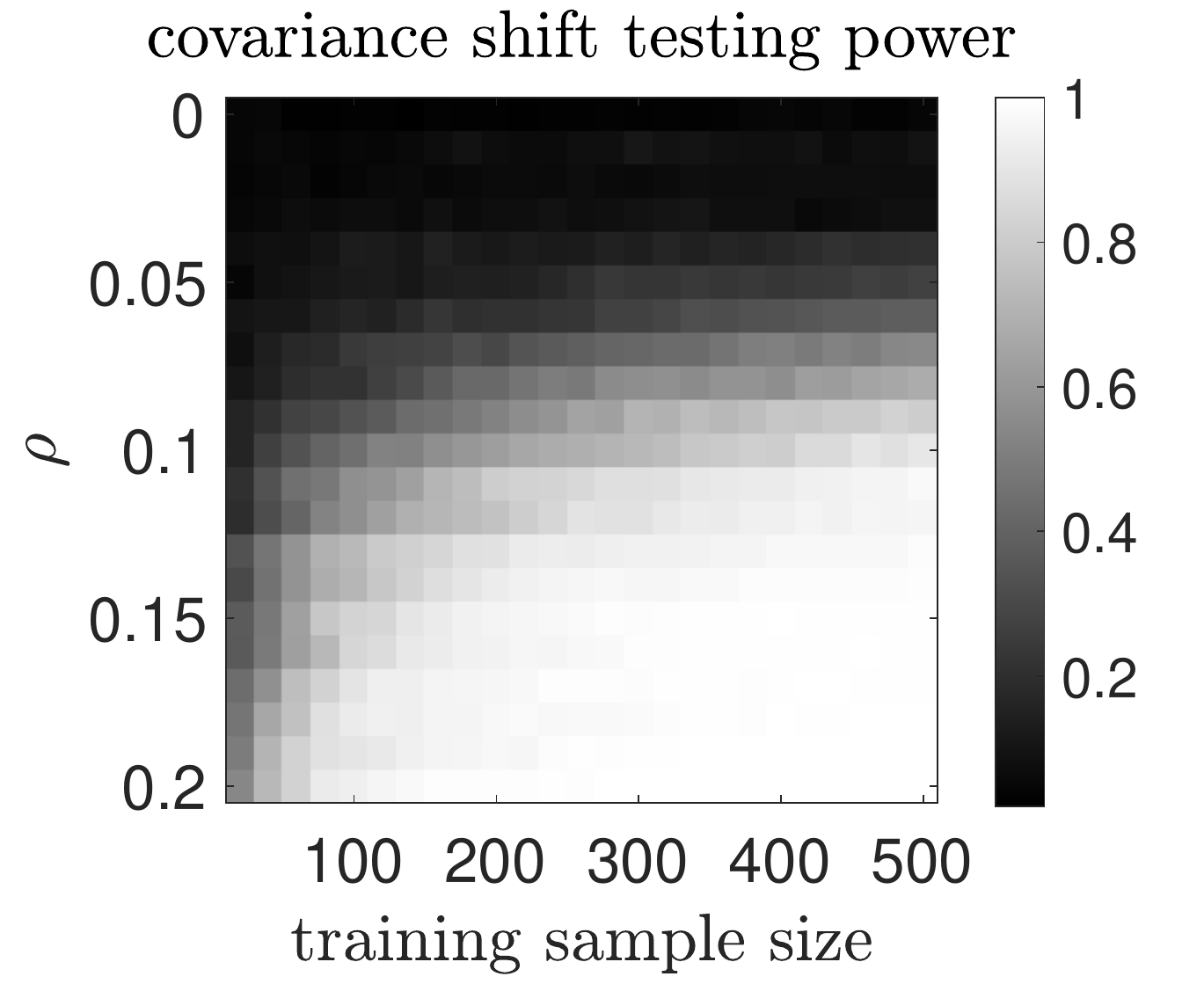} 
\end{center}
\vspace{-8pt}
\caption{
\small
(Left two plots)
Estimated testing power 
on  Gaussian mean shift (change size is $\delta$)
and 
Gaussian covariance shift (change size is $\rho$)
in $\R^{100}$,
where datasets $X$  and $Y$ have 200 samples respectively,
and the training and testing splitting is half-half, i.e. $n_{tr}=n_{te} =200$.
Test power is estimated from $n_{run} = 500$.
(Right two plots) 
Estimated testing power as a function of the number of samples processed in the 1-pass of the training set (batch size =1)
and over varying values of $\rho$,
also plotted as a color field.
}
\label{fig:exp-gaussian}
\end{figure}

{\bf Results}.
The results are shown in the left two plots in Figure \ref{fig:exp-gaussian}. 
The NTK MMD test gives comparable but slightly worse power than the other two benchmarks on the mean shift.
On the covariance shift test,
the network MMD test gives equally good power as the Gaussian MMD.
For the Gaussian covariance shift case,
we also compute the testing power when only part of the training samples are used in the online training,
and the results are shown in the right two plots in Figure \ref{fig:exp-gaussian}.
Testing power increases as the neural network scans more training samples,
\rev{and  when the covariance shifts are larger the transition takes place with smaller training sample size. 
}

In addition, we show in Appendix \ref{appsub:exp-additional}
that NTK-MMD gives similear performance  with varying network architectures, activation functions (like relu), and SGD configurations, and possibly better testing power with a larger network depth and width
(Tables \ref{tab:exp-more-relu} and \ref{tab:exp-more-sgd}).
We also compare  with linear-time kernel MMD in Appendix \ref{appsub:exp-linear-time-MMD}.
As shown in Table \ref{tab:mnist-linear-mmd}, NTK-MMD outperforms linear-time gMMD as in \cite[Section 6]{gretton2012kernel}, 
and underperforms the full gMMD which however requires $O(n^2)$ computation and storage.

\begin{figure}
\begin{center}
\includegraphics[trim =  0 0 0 0, clip, height=.21\linewidth]{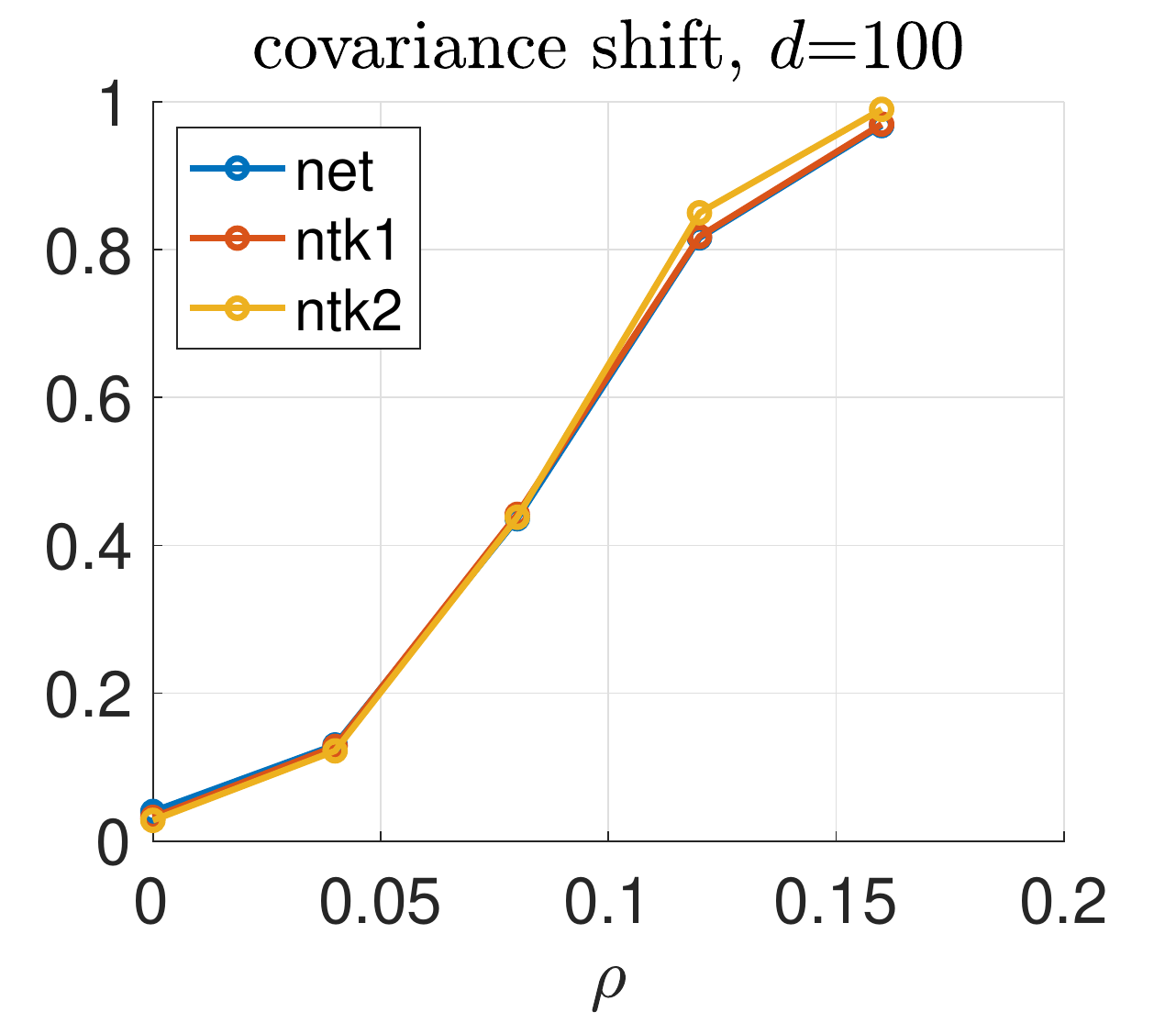} 
\includegraphics[trim =  0 0 0 0, clip, height=.21\linewidth]{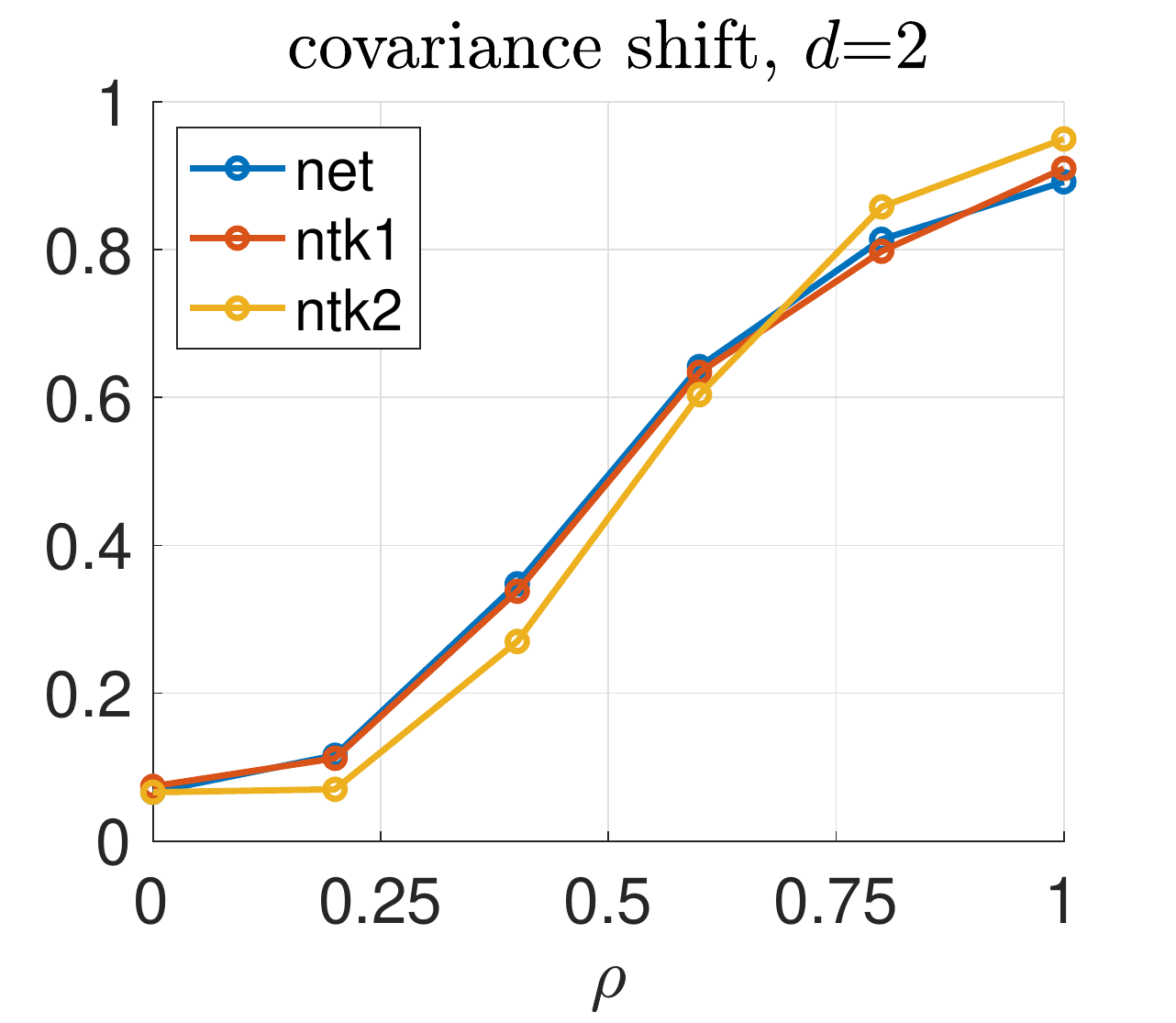} 
\includegraphics[trim =  0 0 0 0, clip, height=.21\linewidth]{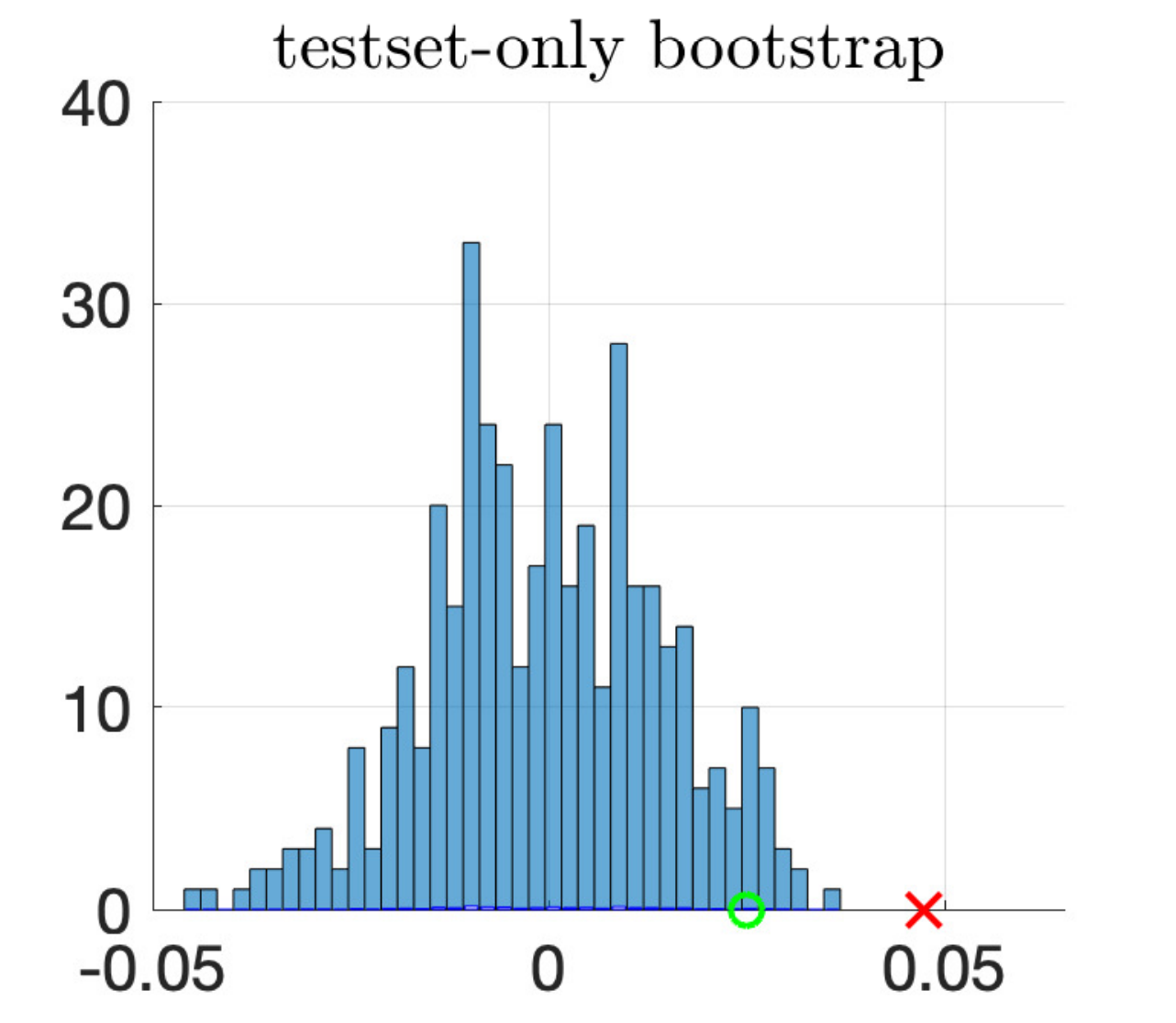}
\includegraphics[trim =  0 0 0 0, clip, height=.21\linewidth]{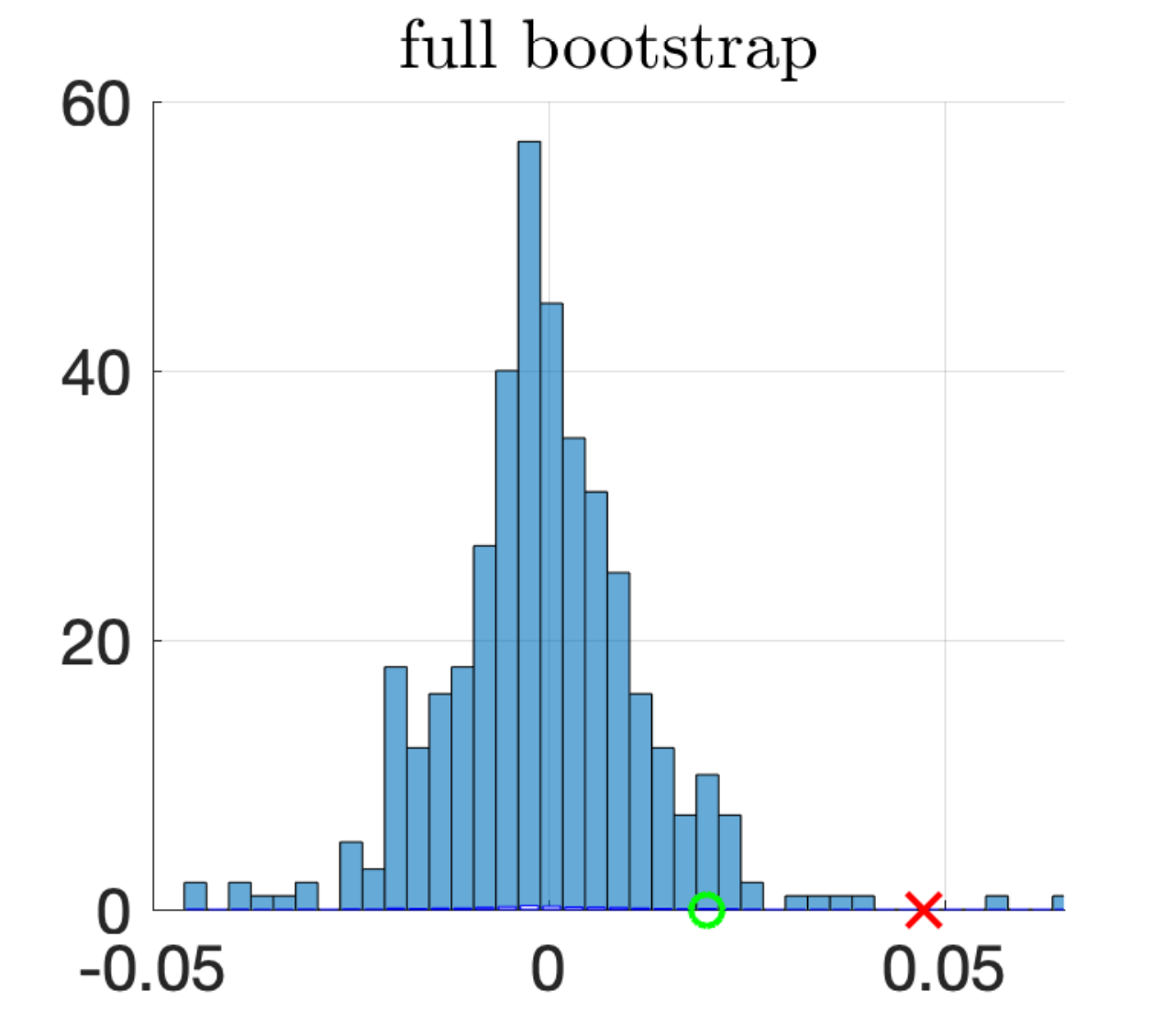} 
\end{center}
\vspace{-11pt}
\caption{
\small
(Left two plots)
Estimated testing power from  $n_{\rm run} = 500 $ of the covariance shift test in Figure \ref{fig:exp-gaussian} in $\R^{100}$
 and $\R^{2}$. 
 $n_X = n_Y = 200$, 
 using three statistics: $\hat{T}_{\net}$ (net), 
 $\hat{T}_{\NTK}$ with test set only bootstrap (ntk1)
 and with full bootstrap (ntk2)
 the training and testing splitting is half-half.
(Right two plots) 
Test statistics $\hat{T}_a$ (red cross), the empirical distribution of $\hat{T}_{a, \rm null}$
using the test-only bootstrap and the full bootstrap (blue bars),
and the estimated threshold (green circle). 
Computed from NTK kernel at $t=0$ and $n_{\rm boot}=400$.
}
\label{fig:exp-ntk}
\vspace{-5pt}
\end{figure}

\subsection{Comparison of $\hat{T}_{\net}$ and $\hat{T}_{\NTK}$}\label{subsec:exp-ntk}

{\bf Set-up}. Since we use a 2-layer fully-connected network,
the finite-width NTK at $t=0$ (using initialized neural network parameters)
can be analytically computed, 
which gives an $n_{te}$-by-$n_{tr}$ asymmetric kernel matrix $K$. 
The expression of $K$ and more details are \rev{provided in Appendix \ref{appsub:exp-exact-more}}. 
This allows computing the exact NTK MMD \eqref{eq:hatT-a},
as well as the
(i) full bootstrap and (i) the test-only  bootstrap 
of the MMD statistic under $H_0$ by 
(i)  permuting  both rows and columns simultaneously 
and (ii) only permuting rows of the matrix $K$. 

{\bf Results.} 
\rev{
To verify the $O(t)$ discrepancy as in Proposition \ref{prop:ntk-O(t)},
we first compute the numerical values of $\hat{T}_{\NTK}$
and $\hat{T}_{\net}(t)$ for different values of $t$
(which corresponds to different learning rate $\alpha$ 
as explained in Remark \ref{rk:SGD} and Appendix \ref{subsec:sgd-bound})
and the relative approximation error defined as $ {\rm err}=|\hat{T}_{\rm net}(t)-\hat{T}_{\rm NTK}|/|\hat{T}_{\rm NTK}|$.
 The results are shown in 
 Figure \ref{tab:O(t)-error}.
The fitted scaling of the error for softplus activation is about $ {0.96}$, 
which agrees with the theoretical $O(t)$ error.  
Switching to relu, the order is not close to 1 (instead ${0.62}$) 
but $\hat{T}_{\rm net}(t)$ still gives a good approximation of $\hat{T}_{\rm NTK}$ as the relative error achieves about $10^{-3}$. 
}
 The comparison of the \rev{testing power} of 
 network approximate NTK statistic 
 and the exact NTK statistic tests 
 are shown in Figure \ref{fig:exp-ntk}.
In the high dimensional Gaussian covariance shift test ($d=100$),
the powers of the three tests are similar.
When reducing dimension to $d=2$,
the full-bootstrap NTK tests show slightly different testing power than the other two.
The network approximate NTK and NTK with test-only bootstrap
always show almost the same testing power,
consistent with the theory in Subsection \ref{subsec:ntk-approx-error}.
In the experiment on $\R^2$ data,
the estimated threshold by full-bootstrap is smaller than by test-only bootstrap (right two plots),
which explains the possibly better-testing power.

\begin{figure}
\begin{center}
 \begin{minipage}[t]{1.05\linewidth}
 \hspace{-20pt}
\includegraphics[trim =  0 0 0 0, clip, height=.165\linewidth]{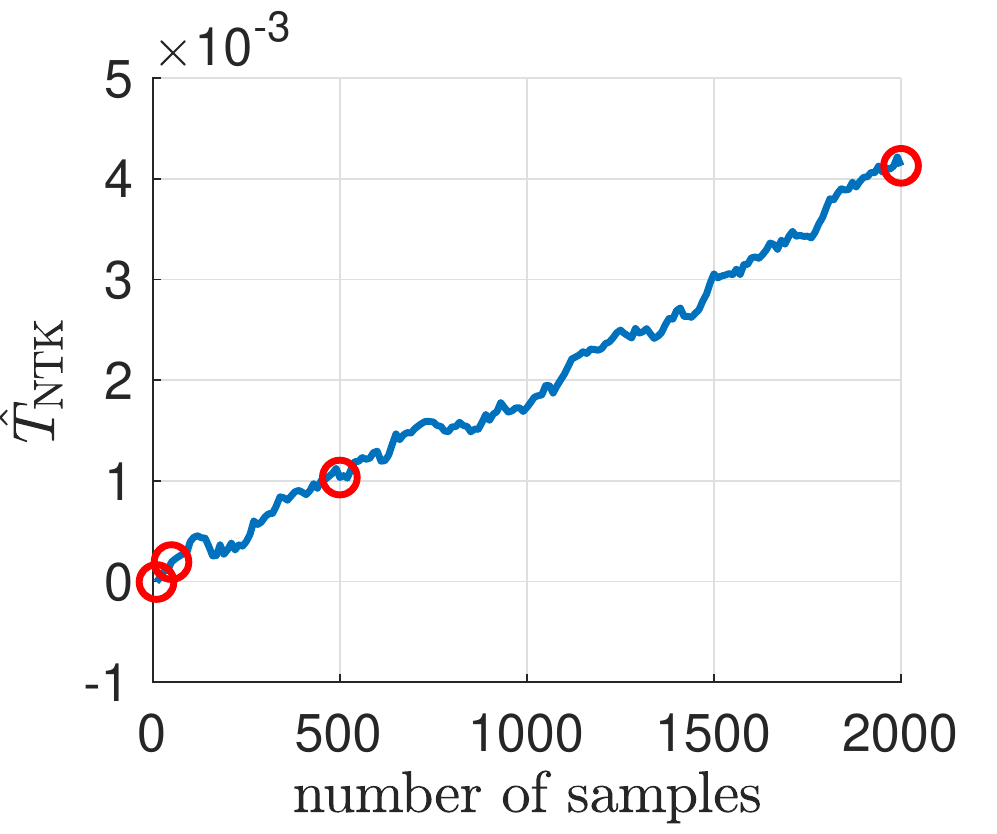} 
\includegraphics[trim =  0 0 0 0, clip, height=.165\linewidth]{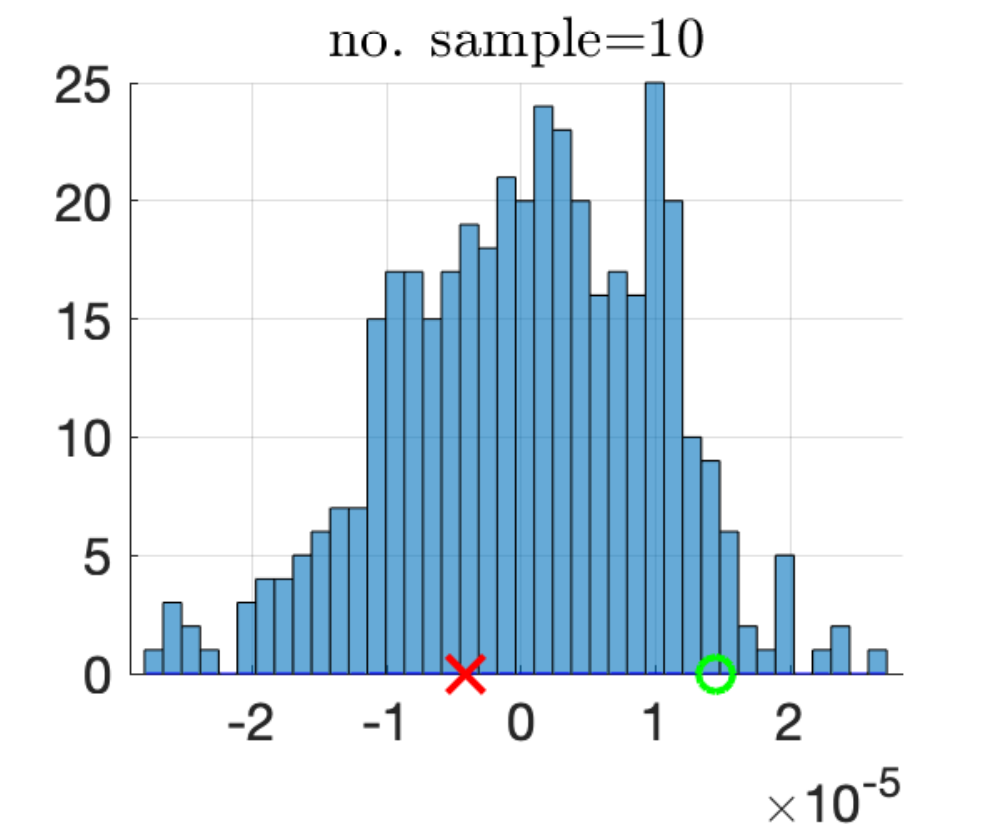} 
\includegraphics[trim =  0 0 0 0, clip, height=.165\linewidth]{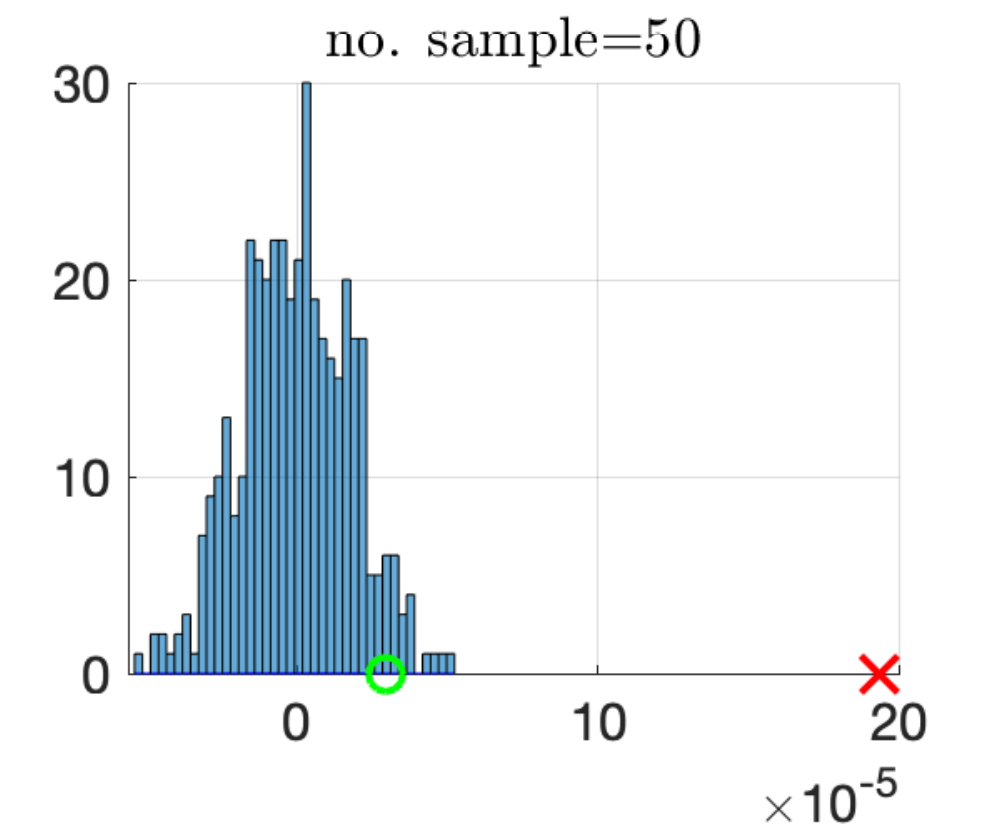} 
\includegraphics[trim =  0 0 0 0, clip, height=.165\linewidth]{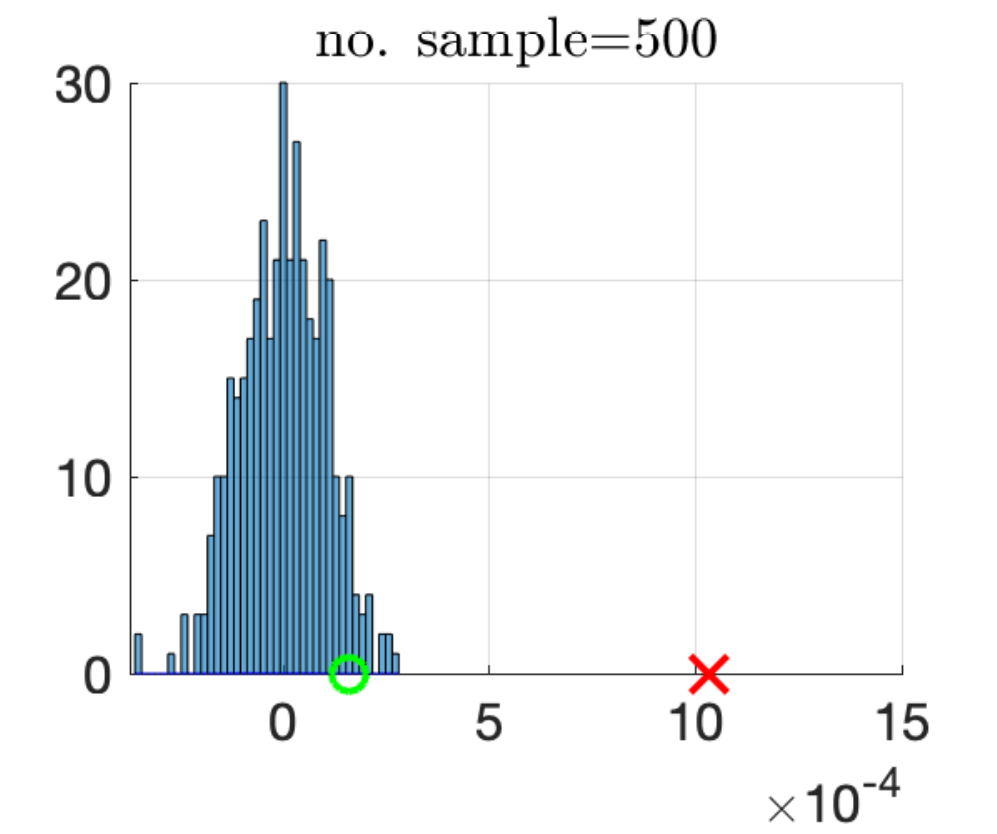} 
\includegraphics[trim =  0 0 0 0, clip, height=.165\linewidth]{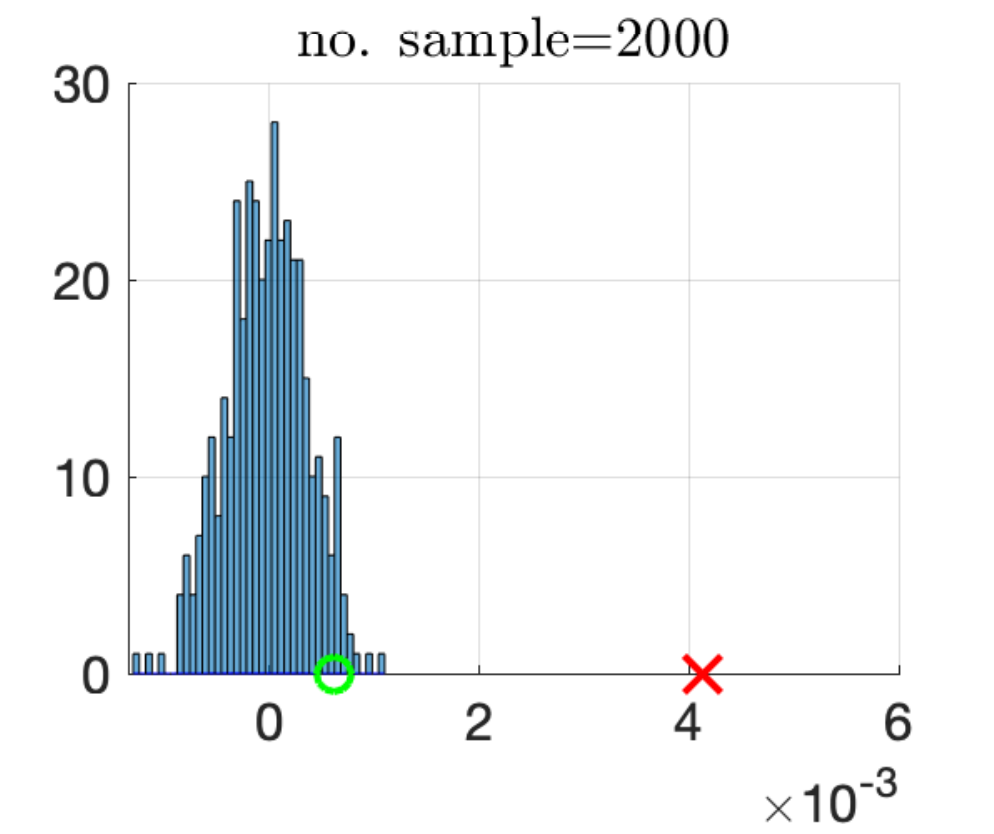} 
\end{minipage}
 \begin{minipage}[t]{1.05\linewidth}
\hspace{-20pt}
\includegraphics[trim =  0 0 0 0, clip, height=.165\linewidth]{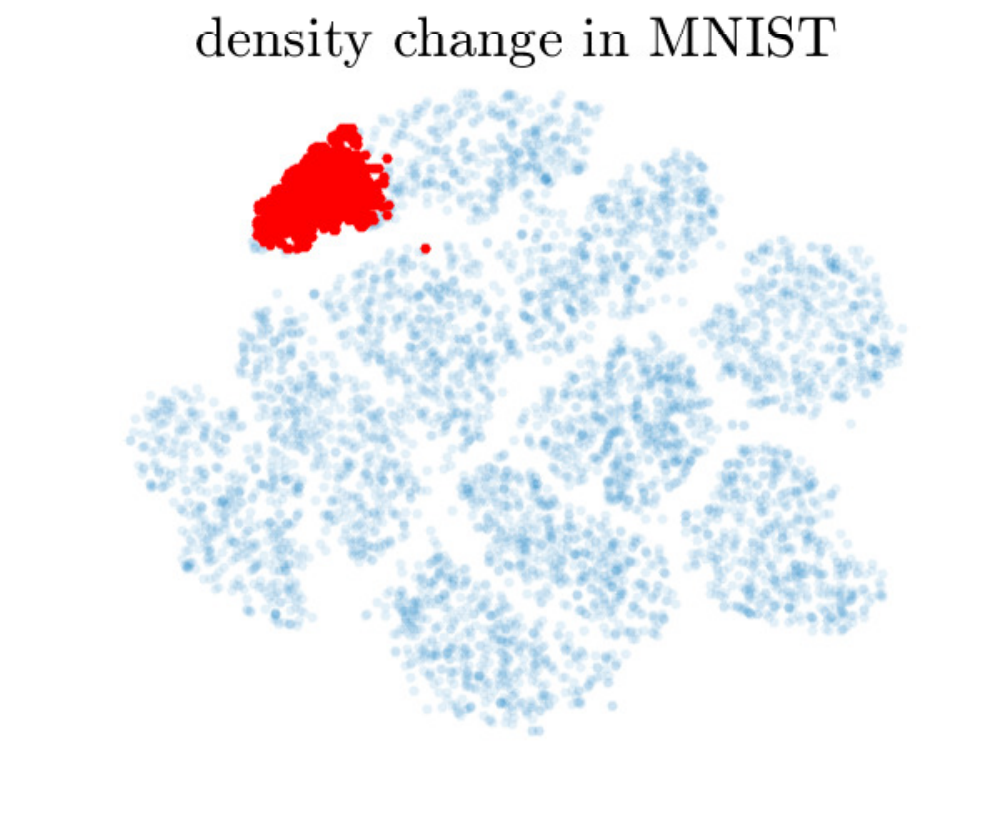} 
\includegraphics[trim =  0 0 0 0, clip, height=.165\linewidth]{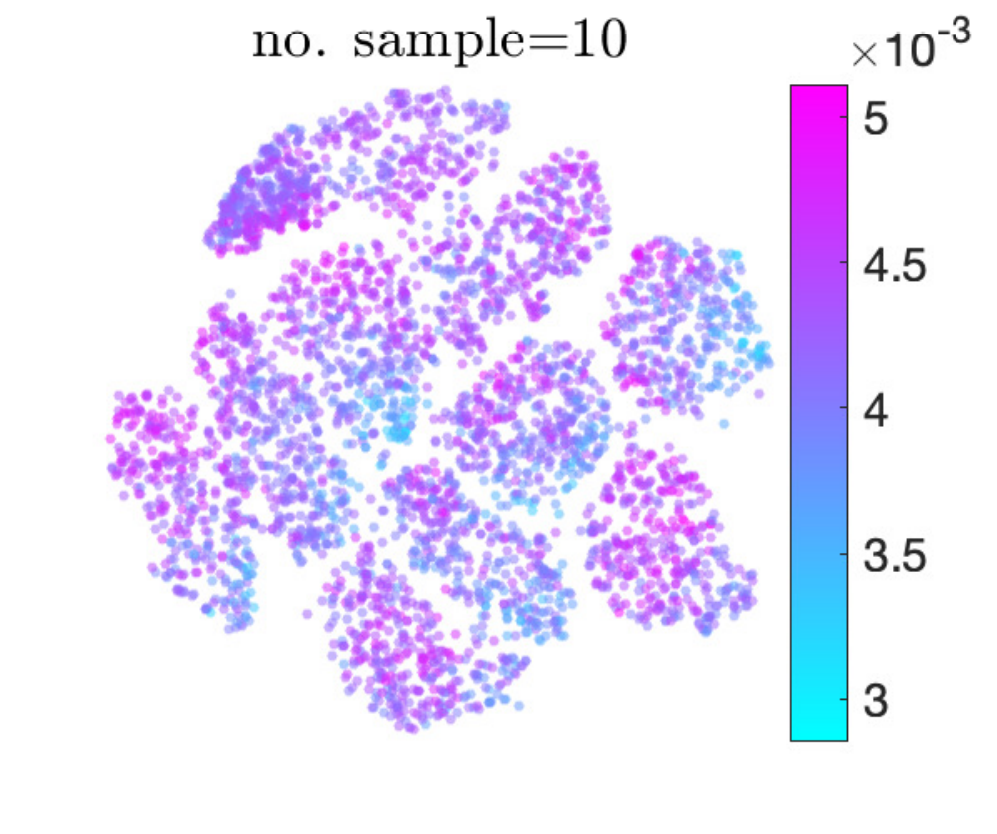} 
\includegraphics[trim =  0 0 0 0, clip, height=.165\linewidth]{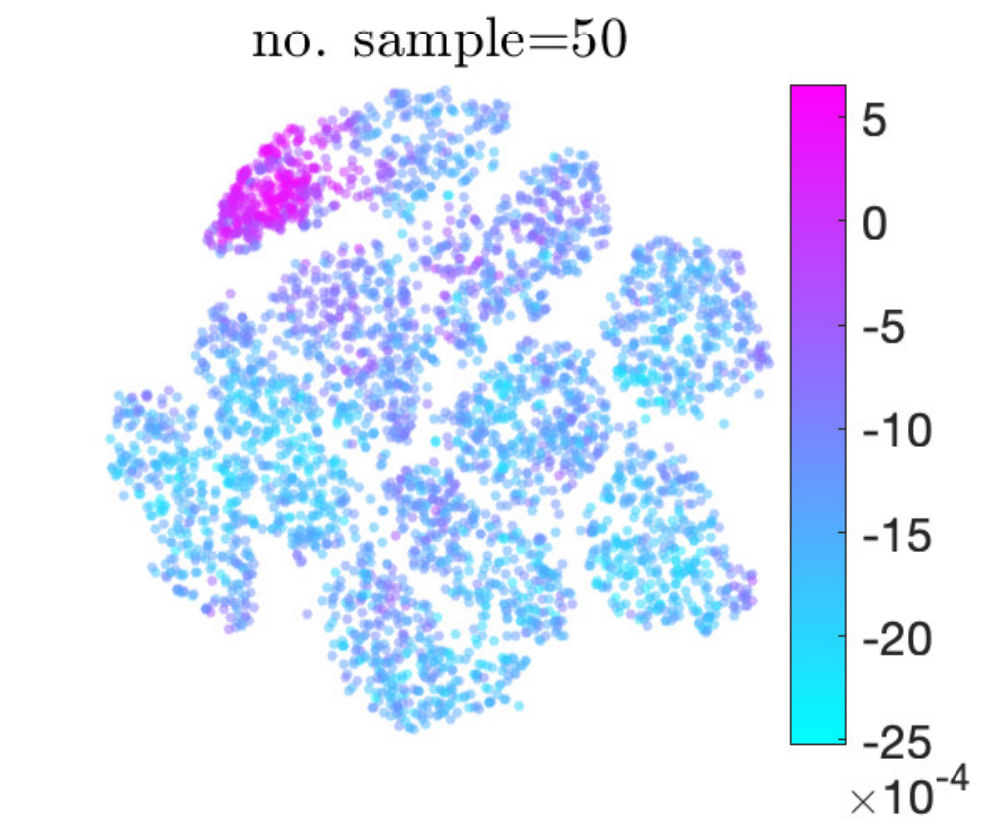} 
\includegraphics[trim =  0 0 0 0, clip, height=.165\linewidth]{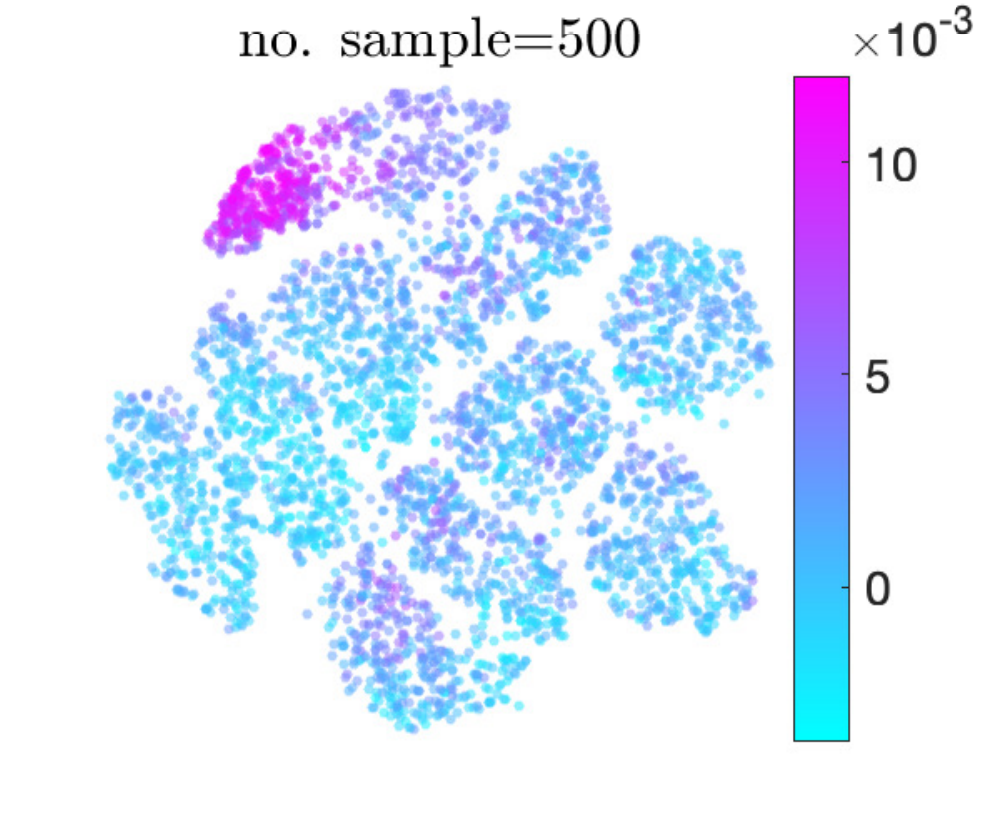} 
\includegraphics[trim =  0 0 0 0, clip, height=.165\linewidth]{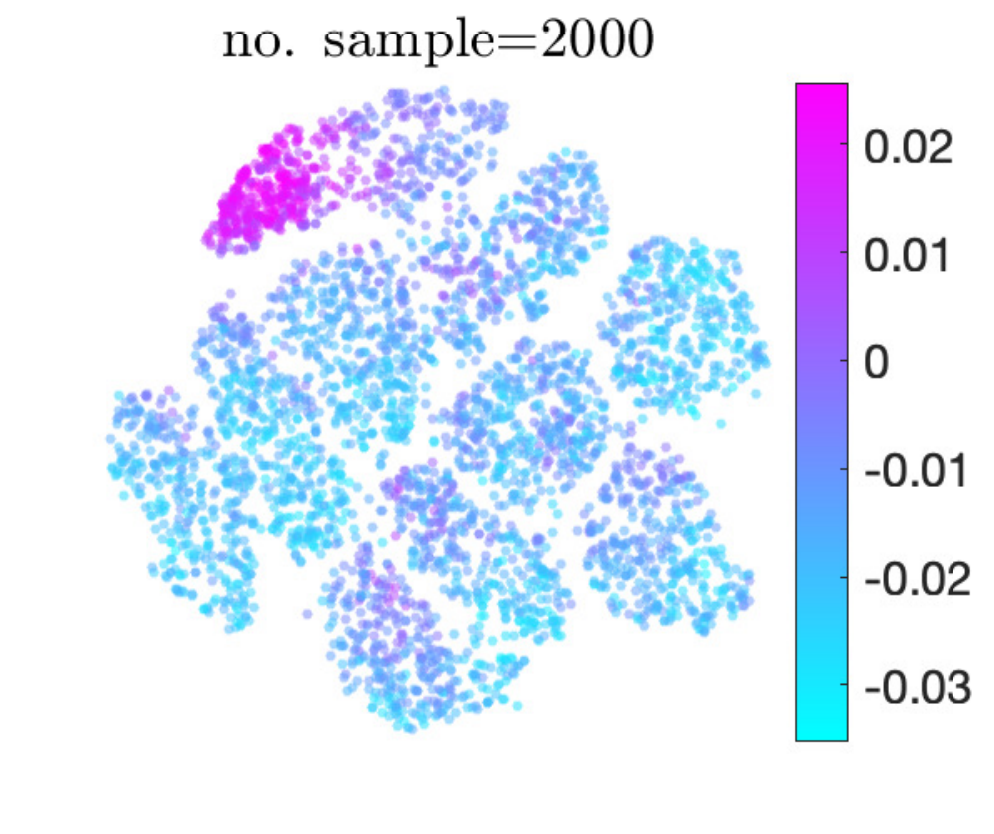} 
\end{minipage}
\end{center}
\vspace{-10pt}
\caption{
\small
NTK-MMD statistic to detect distribution abundance change in MNIST digit image data;
$n_{tr}=2000$ and $n{te} $ is about 4000.
(Top)
Most left: Change of the statistic over the number of samples in the online training (batch size =1) of a 2-layer convolutional network. 
From 2nd-5th columns: The MMD statistic $\hat{T}_a$ compared with the empirical distribution under $H_0$
via test-only bootstrap, at four times along the online training (red circles on the left plot).
(Bottom)
Most left: The change of distribution of MNIST dataset embedded in 2D by tSNE \cite{van2014accelerating}.
From 2nd-5th columns: 
The witness function $\hat{g}$ plotted as a color field over the samples,
at the four times corresponding to the upper panel plots. 
}\vspace{-5pt}
\label{fig:exp-mnist}
\end{figure}

\subsection{Comparison to neural network classification two-sample tests}\label{subsec:exp-C2ST}

{\bf Set-up}. 
We experimentally compare NTK-MMD and state-of-the-art classification two-sample test (C2ST) baselines, which are neural network  based tests. 
Following \cite{liu2020learning}, we compare with C2ST-S, the classification accuracy test  \cite{lopez2017revisiting},
and C2ST-L, the classification logit test \cite{cheng2019classification}. 
Experimental details are given in Appendix \ref{appsub:exp-C2ST-more}.
The data distributions are:

\vspace{5pt}
$\bullet$ Example 1: Gaussian mixture, fixed dimension $d =10$ and increasing $n_{tr}$,
which is the same setting as Figure 3 (left 2 plots) in \cite{liu2020learning}. 
Numbers in Table \ref{tab:gauss-mix-1} show testing power (in \%).  

\vspace{2pt}
$\bullet$ Example 2: Modified Gaussian mixture (from Example 1), the covariance shift is $I+0.1E$ in both mixtures, where $E$ is all-one matrix with zeros on the diagonal. Dimension $d=10$, and number of training samples $n_{tr}$ increases.
The test power is shown in Table \ref{tab:gauss-mix-2}.
\vspace{5pt}

\begin{table}
\centering
\small
\begin{tabular}{l|c c c c}
\hline
~~~~$n_{tr}$	&        2000         &        4000       &       6000       &      	8000    \\          
\hline
ME*             &   $\sim$ 10.0     &      $\sim$ 30.0      &   $\sim$ 58.0          & 	$\sim$ 75.0   \\
SCF*            &       $\sim$ 5.0      &      $\sim$ 6.0   	&   $\sim$ 10.0        &	$\sim$ 15.0   \\
C2ST-S (Adam) 	&    	9.9 (61.6)	  &  14.0 (95.8)   	&   39.1 (100.0)	 &   61.2 (100.0) \\
C2ST-L  (Adam)  &	    14.1 (87.8)	  &  38.4 (100.0)   &	76.4 (100.0) 	 &   92.9 (100.0)   \\		
C2ST-S (SGD)    &   	6.0 (13.9)	  &  10.6 (50.0)    &	10.8 (94.4) 	 &   14.8 (99.6)    \\
C2ST-L  (SGD)   &   	6.7 (22.2)	  &  12.8 (81.6)    &	22.1 (100.0)	 &   34.6 (100.0)   \\
NTK-MMD         &         7.1 		  &      9.6 		&    13.7 		    &      17.9         \\
\hline
\end{tabular}
\vspace{-5pt}
\caption{
Test power on Gaussian mixture data Example 1, dimension $d=10$.
(*recovered from Figure 3 in \cite{liu2020learning}, $\sim$ means about.) 
For C2ST’s, the number outside brackets is for epoch $=1$, and in brackets for epoch $=10$.
}\label{tab:gauss-mix-1}
\end{table}

\begin{table}
\centering
\small
\begin{tabular}{l|c c c c}
\hline
~~~~~  $n_{tr}$     &   500            &          1000         &	  	1500          &      2000  \\                 
\hline
C2ST-S (Adam)      	 &     	21.8 (28.1)	&    62.2 (53.8) 	    &   79.4 (74.0)   &	94.6 (85.2) \\
C2ST-L  (Adam)      &     	48.5 (49.4) &	 92.8 (82.6) 	    &   99.5 (96.3)	  &	100.0 (98.8)\\
C2ST-S (SGD)        &     	7.4 (28.3) 	&    22.7 (79.7) 	    &   35.3 (92.4)   &	 54.9 (96.8) \\
C2ST-L  (SGD)       &      18.3 (52.2) 	&    56.8 (97.6)  	    &   81.4 (99.9)   &	 97.3 (100.0)\\
NTK-MMD      		&     	34.3 	 	&    68.9 	         	&   88.8  		  &  95.9  \\
\hline
\end{tabular}
\vspace{-5pt}
\caption{
Test power on Gaussian mixture data Example 2, dimension $d=10$.
}\label{tab:gauss-mix-2}
\end{table}

{\bf Results}. 
On Example 1, NTK-MMD performs similar to SCF test in most cases, better than C2ST-S (SGD 1-epoch), and is worse than the other baselines. On Example 2, NTK-MMD outperforms C2ST baselines in several cases, e.g., constantly better than C2ST-S (SGD and Adam, 1-epoch) and comparable to C2ST-L (SGD 1-epoch).
Note that C2ST baselines can be sensitive to training hyperparameters, such as the choice of optimization algorithm (SGD or Adam) and number of epochs. As far as the authors are aware of, there is no theoretical training guarantee of C2ST tests. In contrast, NTK-MMD has theoretical training guarantees due to the provable approximation to a kernel MMD. The weakness of NTK-MMD, though, is that the NTK kernel may not be discriminative to distinguish certain distribution departures, like in Example 1. The expressiveness power of NTK-MMD may be theoretically analyzed, for example, in the infinite-width limit using the analytical formula, as the infinite-width NTK has been shown to be universal for data on hyperspheres \cite{jacot2018neural}.
Overall, the results suggest that the performances of the three neural network tests depend on the data distributions, which is anticipated for any hypothesis test. 
Further theoretical investigations are postponed here.

\subsection{MNIST distribution abundance change}\label{subsec:exp-mnist}

{\bf Dataset}. 
We take the original MNIST dataset, which contains  28$\times$28 gray-scale images,
and construct two densities $p$ and $q$ by subsampling from the 70000 images in 10 classes, following \cite{ChengXie2021}:
$p$ is uniformly subsampled from the MNIST dataset, $p=p_{\rm data}$,
and $q$ has a change of abundance
$ q = 0.85 p_{\rm data} + 0.15 p_{\rm cohort}$, 
where 
$p_{\rm cohort}$ is the distribution of a subset of the class of digit ``1'' having about 1900 samples.
The $p_{\rm cohort}$ is illustrated in the left bottom plot in Figure \ref{fig:exp-mnist}.
The two samples $X$ and $Y$ have $n_X = 3000$, $n_Y=2981$,
and we randomly split $X$ and $Y$ make the training set $\calD_{tr}= \{X_{(1)}, Y_{(1)}\}$, $n_{X,(1)}= n_{Y,(1)}=1000$, and the rest is the test set $\calD_{te}$. 

{\bf Results}.
Using a 2-layer convolutional nerual network,
we compute the network MMD statistic $\hat{T}_{a,\rm net}$ \eqref{eq:hatT-a-net} and the test-only bootstrap \eqref{eq:def-T-null-split}.
 The online training uses batch size =1 and one epoch, and more experimental details are in \rev{Appendix \ref{appsub:exp-mnist-more}}. 
The results are shown in Figure \ref{fig:exp-mnist}. 
The NTK-MMD statistic already shows testing power after being trained on 50 samples, 
and in the later stage of training,
the NTK witness function $\hat{g}_{(1)}$
identifies the region of the abundance change.

\subsection{Online human activity change-point detection}

\begin{figure}[t]
\begin{minipage}[t]{1.05\linewidth}
\hspace{-23pt}
\raisebox{10pt}{
\includegraphics[trim =  0 0 0 0, clip, height=.18\linewidth]{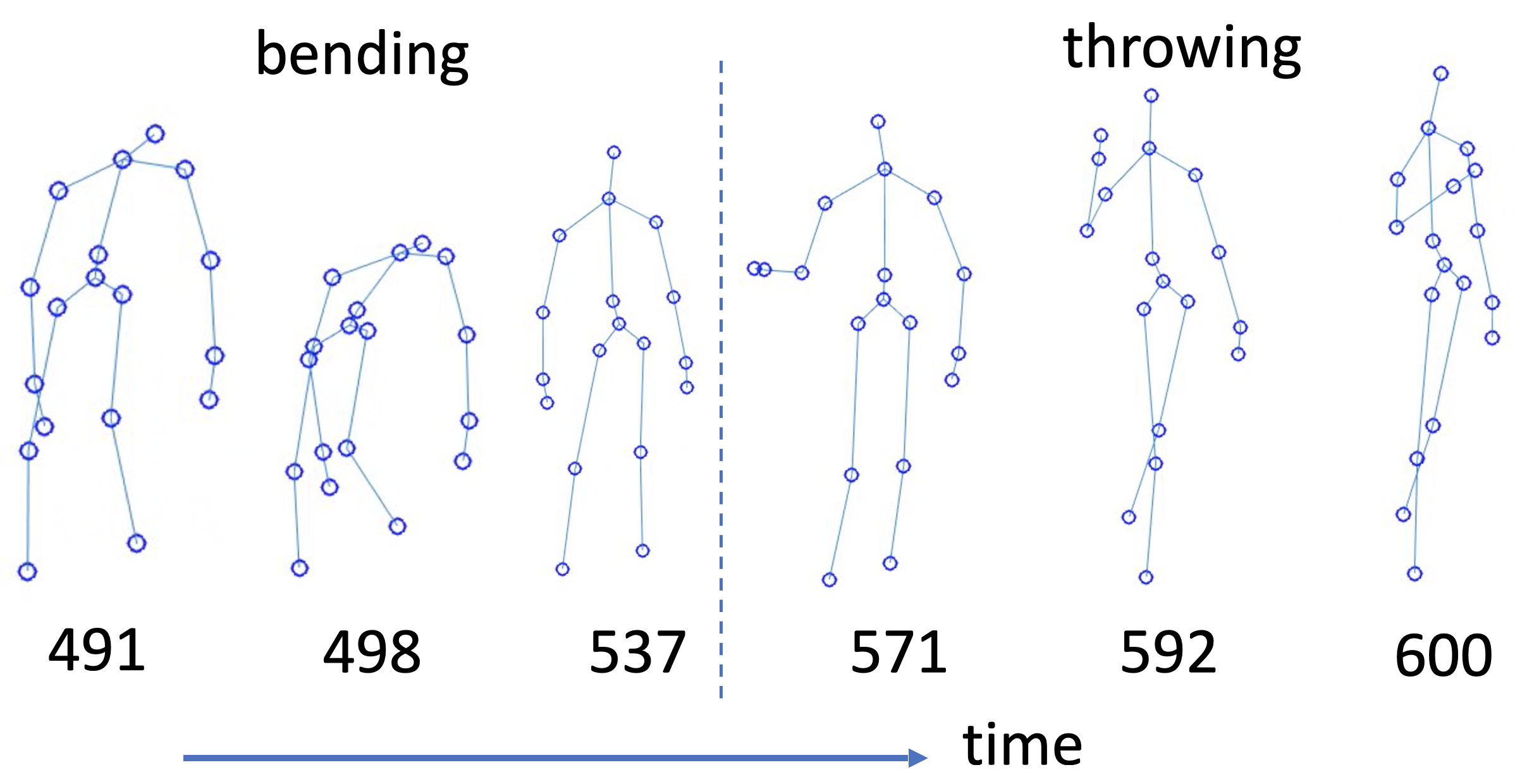} }
\hspace{3pt}
\includegraphics[trim =  0 0 0 0, clip, height=.2\linewidth]{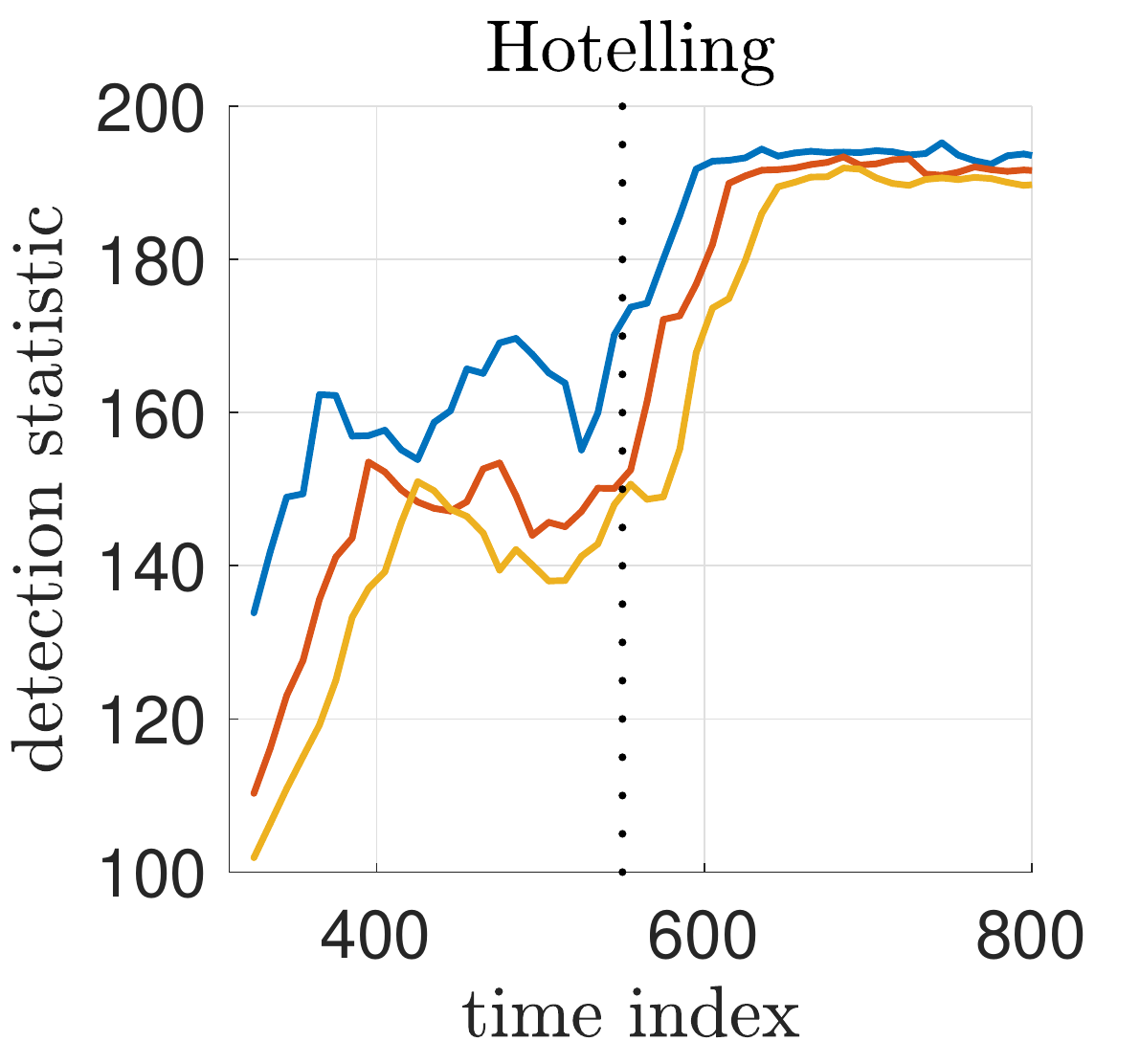} 
\includegraphics[trim =  0 0 0 0, clip, height=.2\linewidth]{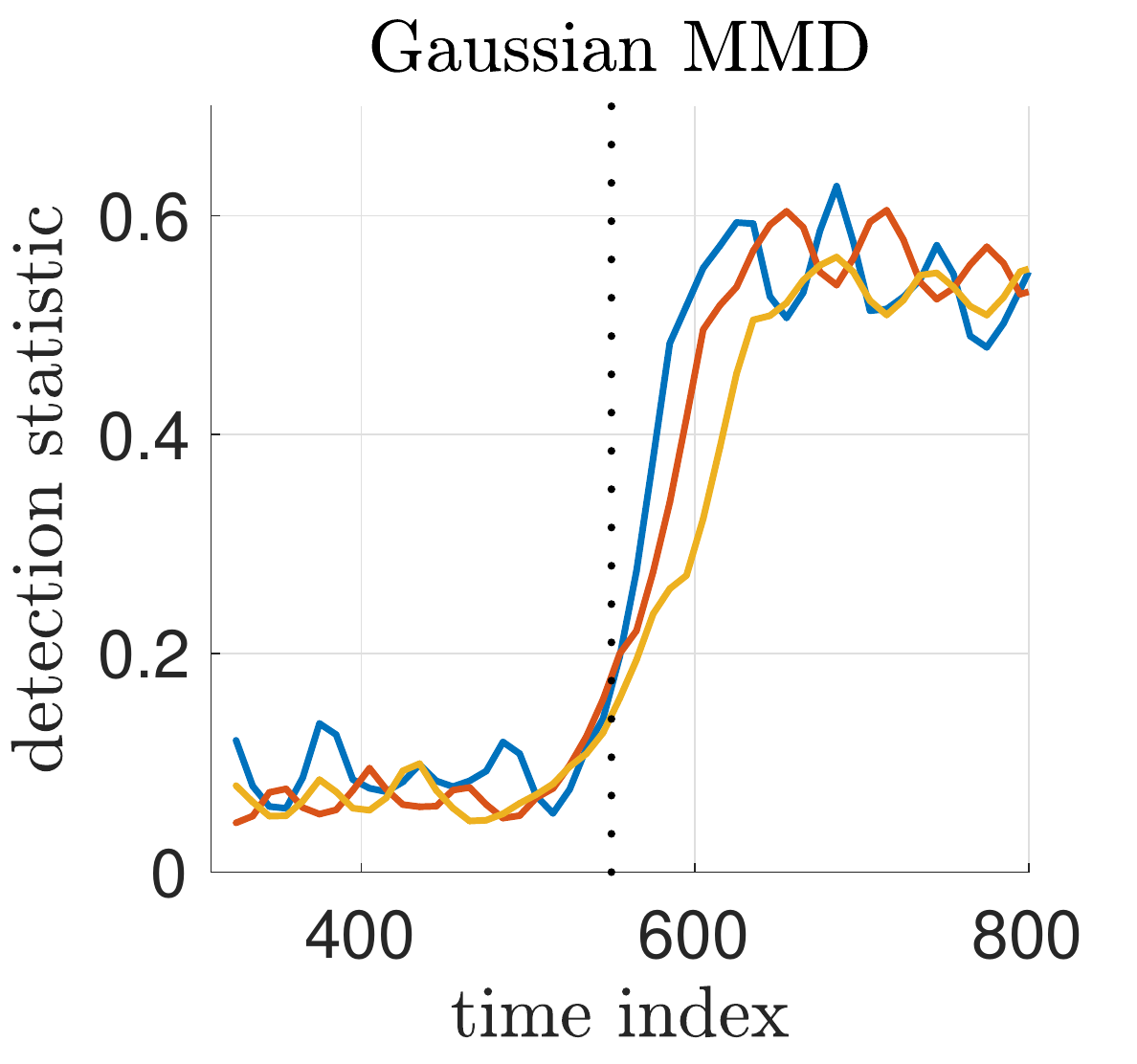} 
\includegraphics[trim =  0 0 0 0, clip, height=.2\linewidth]{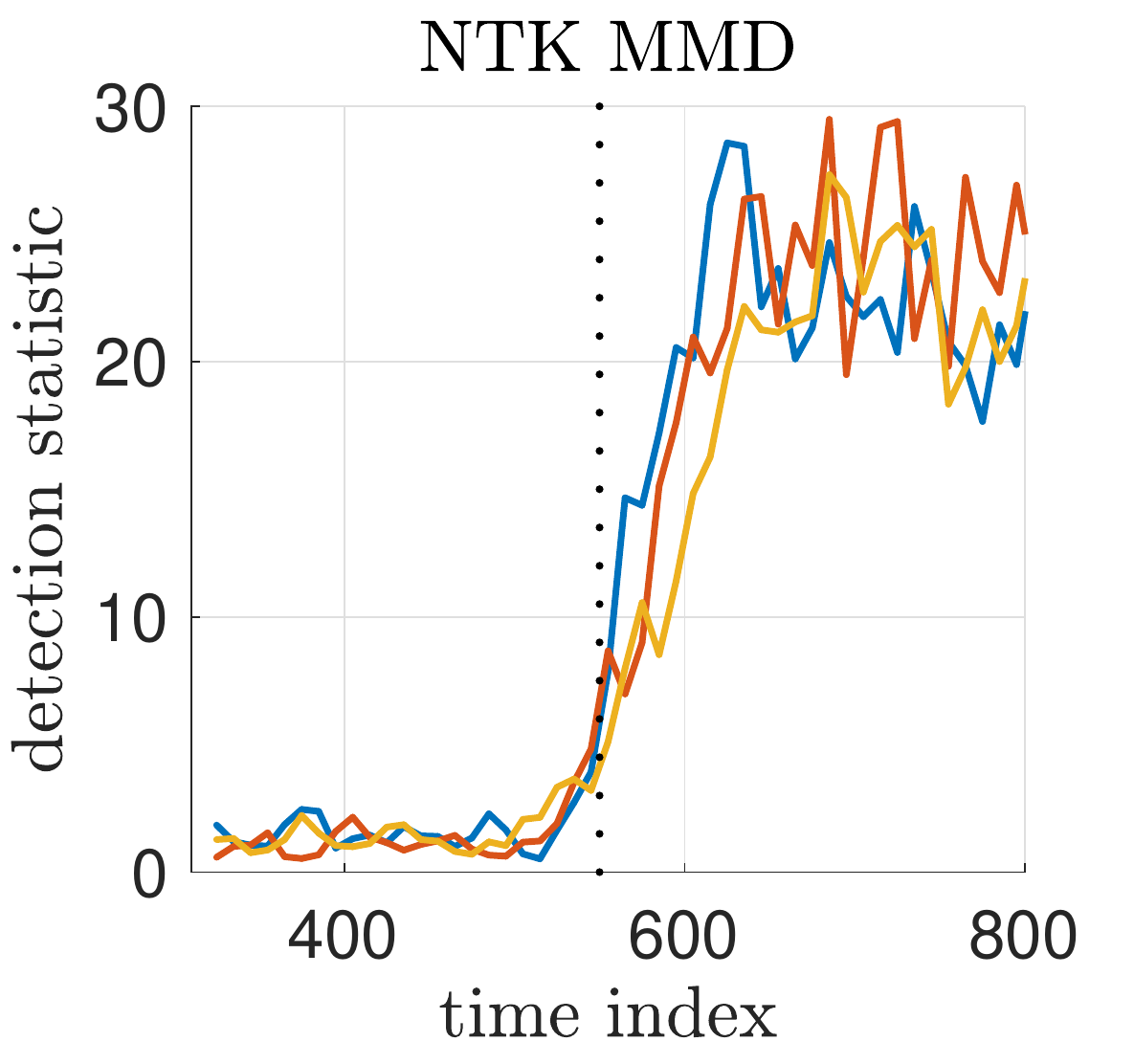} 
\end{minipage}
\vspace{-15pt}
\caption{
\small
(Left) Example data sequence in the human activity dataset: 
before and after the change-point.
(Right)
Detection statistics computed by Hotelling's T statistic,
Gaussian MMD,
and NTK MMD,
on human action trajectory dataset, using window size 100 (blue), 150 (red), and 200 (yellow), respectively.
The change point is at time index 550 (dotted black).
}\vspace{-5pt}
\label{fig:exp-action}
\end{figure}

{\bf Set-up}. 
We present an illustrative example using NTK-MMD test statistic to perform online change-point detection: 
detecting human activity transition.
We consider a real-world dataset, the Microsoft Research Cambridge-12 (MSRC-12) Kinect gesture dataset \cite{fothergill2012instructing}. 
\rev{
The data sequence records a human subject repetitively bending the body/picking up and throwing a ball
before/after the change happens.}
After preprocessing, the sequential dataset contains
1192 frames (samples) and 54 attributes (data samples are in $\R^{54}$),
with a change of action from ``bending'' to ``throwing'' at time index 550.
More description of the dataset and experimental details is provided in \rev{Appendix \ref{appsub:exp-change-point-more}}.
Example samples before and after the change point are shown in the left of Figure \ref{fig:exp-action}. 
The algorithm is based on a sliding window which moves 
forward with time, and we compute the detection statistic 
every ten frames; 
such a procedure can be viewed as the Shewhart Chart in the literature \cite{xie2021sequential}; scanning MMD statistic has been used in \cite{li2019scan}. 
The window size is chosen to be 100, 150, and 200, respectively. We take a block of data (same size as the window) before the time index 300 (to use as the pilot samples) and compare with the distribution of data from the sliding window
to compute the detection statistic. If there is a change-point, 
the detection statistic will show a large value. 

{\bf Results}.
The other two detection statistics are computed by  (i) Gaussian MMD (with bandwidth chosen to be median distance) and (ii) Hotelling's T statistics.
The results are shown in Figure \ref{fig:exp-action},
where both the Gaussian MMD and the NTK-MMD statistics can detect the change: the detection statistic value remains low before the change and remains high after the change point, and both  are better than the Hotelling statistic.

\section{Discussion}\label{sec:discussion}

The current work can naturally be extended in several aspects. 
First, the analysis of NTK approximation error may be extended,
e.g., to other activation functions, and under the infinite-width limit.
Second, considering other training objectives may allow us to compare NTK-MMD to other neural network classification tests.
At the same time, 
the limitation of lazy-regime training has been studied in \cite{ghorbani2019limitations,ghorbani2020neural,refinetti2021classifying}, which indicates that NTK theory cannot fully characterize the modeling ability of deep networks.  
It has also been shown that the expressiveness of the NTK kernel may be restricted to certain \rev{limited type of} kernels
\cite{bietti2019inductive,geifman2020similarity,chen2020deep}.
This motivates extensions of NTK for studying deep network training \cite{huang2020dynamics,roberts2021principles}.
Finally, the application may  extend to various hypothesis testing tasks as well as deep generative models.
We thus view the current work as a first step towards understanding the role and potential of trained neural networks in testing problems and applications. 


\section*{Acknowledgement}
The authors thank 
Alexander Cloninger
and
Galen Reeves
for helpful discussion on the initial version of the paper,
and the anonymous reviewers for helpful feedback. 
The work is supported by NSF DMS-2134037.
XC is partially supported by NSF, NIH and the Alfred P. Sloan Foundation.
YX is partially supported by NSF (CAREER Award 1650913, DMS-1938106, and DMS-1830210).

\bibliographystyle{plain}
\bibliography{nn,mmd,kernel}

\appendix

\setcounter{table}{0}
\setcounter{figure}{0}
\setcounter{lemma}{0}
\renewcommand{\thetable}{A.\arabic{table}}
\renewcommand{\thefigure}{A.\arabic{figure}}

\section{Proofs and additional analysis in Section \ref{sec:method}}

\subsection{Proofs \rev{in Section \ref{subsec:ntk-approx-error}}}\label{appsub:proofs-sec2}

\begin{remark}[Biased and unbiased MMD estimators]\label{rk:unbiased-mmd}
\rev{
The exact NTK-MMD statistic \eqref{eq:def-hatT} is a biased estimator \cite{gretton2012kernel}. 
The unbiased estimator is by excluding the diagonal terms of kernel matrix in the summation and normalizing by ``$1/n(n-1)$'' instead of  ``$1/n^2$''. 
We consider biased estimator for simplicity, 
and also because the testing power analysis gives similar results, c.f. the comment beneath Theorem \ref{thm:power}.
In addition, the ``asymmetric MMD statistic'' \eqref{eq:hatT-a} (with training-test splitting and used in many practical situations) is an unbiased estimator.
}
\end{remark}

\begin{proof}[Proof of Lemma \ref{lemma:u(x,t)}]

By \eqref{eq:def-Loss}, 
\begin{equation}
\frac{\partial \hat{L}}{\partial \theta} = 
- \int_{\calX} \nabla_\theta f( x; \theta) (\hat{p} -  \hat{q} )(x) dx.
\end{equation}
By the GD training dynamic, 
\begin{align}
\frac{\partial}{\partial t} u(x,t)
& =
\langle \nabla_\theta f(x, \theta(t)), \dot{\theta}(t) \rangle \\
& =     \left\langle \nabla_\theta f(x, \theta(t)), - \frac{\partial \hat{L}}{\partial \theta} \right\rangle 
\label{eq:partial-t-f-eqn1.1} \\
& = 
 \int_{\calX} \langle \nabla_\theta f( x; \theta(t)), \nabla_\theta f( x'; \theta(t)) \rangle
 (\hat{p} -  \hat{q})(x') dx'.
\label{eq:partial-t-f-eqn1}
\end{align}
This proves the lemma by definition of $\hat{K}_t(x,x') $ in \eqref{eq:def-hatKt}. 
\end{proof}

\begin{proof}[Proof of Lemma \ref{lemma:ntk-approx}]

To prove Part (1):
The initial weights
$\theta(0) \in \Theta$, and by Taylor expansion, 
\[
\theta(t) = \theta(0)+ t \dot{\theta}( t' ),
\]
from $ 0 < t' < t < t_{f,r}$.
By definition, 
\begin{equation}\label{eq:dot-theta-t'}
\dot{\theta}(t')  
=\left.  -\frac{\partial \hat{L}}{\partial \theta} \right |_{\theta = \theta( t')}
= 
 \int_{\calX} \nabla_\theta f( x; \theta(t')) (\hat{p} -  \hat{q})(x) dx,
\end{equation}
and thus by that  $\| \nabla_\theta f  \|_{\calX, \Theta} \le L_f$.
\[
\| \dot{\theta}(t')  \|
\le 
2 L_f,
\]
This  give that $\| \theta(t) - \theta(0) \| \le t 2 L_f$, which proves Part (1). 

To prove Part (2):
Note that  for any $x, x'$, and $t < t_{f,r}$,
\[
\frac{\partial }{ \partial t} \hat{K}_t(x, x')
= 
R_{t}(x,x') + R_{t}(x', x), 
\quad
R_{t}(x,x'):=
\langle {\rm D}_\theta^2 f (x, \theta(t))( \dot{\theta}(t)), \nabla_\theta f( x', \theta(t)) \rangle.
\]
By Taylor expansion, for some $0 < t' < t$, 
\[
\hat{K}_t(x,x') = K_{0}(x,x')  + t ( R_{t'}(x,x') + R_{t'}(x', x) )
\]
where by part (1), $\theta(t') \in B_r$. 
Again by \eqref{eq:dot-theta-t'}, this gives that
\[
\| \dot{\theta}(t')  \|
\le 
2 \| \nabla_\theta f \|_{\calX, B_r},
\]
{where note that the domain of $\theta$ is $B_r \subset \Theta$,
and thus the constant $\| \nabla_\theta f \|_{\calX, B_r}$ can potentially be smaller than $L_f$.}
Then,
\[
\| {\rm D}_\theta^2 f (x, \theta(t'))( \dot{\theta}(t')) \| 
\le  2 \| {\rm D}_\theta^2 f\|_{\calX, B_r} \| \nabla_\theta f\|_{\calX, B_r}.
\]
As a result,
\[
|R_{t'}(x,x') |
\le \| {\rm D}_\theta^2 f (x, \theta(t'))( \dot{\theta}(t')) \|  
\|  \nabla_\theta f( x', \theta(t'))  \|
\le 2 \| {\rm D}_\theta^2 f\|_{\calX, B_r} \| \nabla_\theta f\|_{\calX, B_r}^2.
\]
The same bound holds for $| R_{t'}(x, x') | $,
and the above bounds are uniformly for all $x, x'$. 
Putting together, this proves Part (2). 
\end{proof}

\begin{proof}[Proof of Proposition \ref{prop:ntk-O(t)}]
By definition,
\begin{align}
& \hat{T}_{net}(t)  - \hat{T}_{\rm NTK}
 = \int_\calX \int_\calX  
 \frac{1}{t} \int_0^t  \left(  \hat{K}_s(x,x')  - K_0(x,x')\right) ds 
(\hat{p} - \hat{q})(x') dx'  (\hat{p} - \hat{q})(x) dx  \nonumber  \\
& ~~~
=  \int_\calX \int_\calX  E(x,x')
 \hat{p}(x')\hat{p}(x) dx'dx
 -  \int_\calX \int_\calX  E(x,x')
 \hat{p}(x')\hat{q}(x) dx'dx  \nonumber \\
& ~~~~~~
-  \int_\calX \int_\calX  E(x,x')
 \hat{q}(x')\hat{p}(x) dx'dx
 +  \int_\calX \int_\calX  E(x,x')
 \hat{q}(x')\hat{q}(x) dx'dx 
 \label{eq:pf-prop-error-1}
\end{align}
where we define
\[
E(x,x'):= \frac{1}{t} \int_0^t  \left(  \hat{K}_s(x,x')  - K_0(x,x')\right) ds.
\]
By Lemma \ref{lemma:ntk-approx},  for $t < t_{f,r}$ and for any $x, x' \in \calX$
\[
|E(x,x')| \le 
 \frac{1}{t} \int_0^t  \left|  \hat{K}_s(x,x')  - K_0(x,x')\right| ds
 \le 
  \frac{1}{t} \int_0^t C_{f,r} s ds = \frac{t}{2} C_{f,r}.
\]
Thus the four terms in \eqref{eq:pf-prop-error-1} 
in absolute value are all upper bounded by $ C_{f,r} t/2$,
and thus
$| \hat{T}_{net}(t)  - \hat{T}_{\rm NTK}|$ is upper bounded by the sum of the absolute values of the four terms
which is less than or equal to $2 C_{f,r} t$.
\end{proof}

\subsection{Extension to SGD training}\label{subsec:sgd-bound}

Consider the online setting of training the network by minimizing the loss $\hat{L}(\theta)$ in \eqref{eq:def-Loss}
on $n=n_X+ n_Y$ samples.
We write the training set $\calD_{tr} = \{  (z_i, l_i) \}_{i=1}^n$, where 
$z_i$ is from $X$ or $Y$, and $l_i = 1$ or 2 is the class label. 
Let $b_i = 1/n_X$ if $l_i = 1$, and $1/n_Y$ if $l_i= 2$.
The loss can be written as 
\[
\hat{L}(\theta) = \sum_{i=1}^n f(z_i; \theta) b_i,
\quad b_i = \begin{cases}
-1/n_X, & \quad l_i = 1, \\
 1/n_Y, & \quad l_i = 2.
\end{cases}
\]
For simplicity, assume that $n_X = n_Y = n/2$. We define $l_i(\theta ) = b_i f(z_i ; \theta)$, which is the loss from the $i$-th sample. 

Suppose we train the network with batch size =1 and 1 epoch.
The learning rate is $\alpha$, that is, for $k$-th iteration in the SGD, $k=1,\cdots, n$,
\[
\theta_k = \theta_{k-1} - \alpha \nabla_\theta l_k(\theta_{k-1}), 
\]
from some $\theta_0 \in \Theta$. 
Note that 
$\nabla_\theta l_k ( \theta) = b_k  \nabla_\theta f(z_k; \theta)$, 
and thus
\begin{equation}
\|\theta_k - \theta_{k-1}\| 
= \alpha \| \nabla_\theta l_k ( \theta_{k-1}) \| 
= \alpha |b_k| \|\nabla_\theta f(z_k; \theta_{k-1}) \|
\le  2 L_f \frac{ \alpha }{n} .
\end{equation}
This implies that 
\begin{equation}\label{eq:|thetak-theta0|}
\| \theta_k - \theta_0\| \le 2 L_f  \frac{ k}{n} \alpha,
\end{equation}
and in particular,  $\| \theta_n - \theta_0 \| \le 2L_f \alpha$.
Thus, $\theta_k$ for all $k$ up to $n$ stays in a $r$-Euclidean ball of $\theta_0$ if $2L_f \alpha < r$.

We write the network function at $k$-th step as $u_k$,
$u_k(x) = f(x; \theta_k)$. 
\begin{align}
& u_{k}(x) - u_{k-1}(x) 
 = f(x; \theta_k) - f(x; \theta_{k-1}) \nonumber \\
& ~~~ 
= \nabla_\theta f(x; \theta_{k-1})^T( \theta_k - \theta_{k-1}) 
+ O(  \| \theta_k - \theta_{k-1} \|^2) \nonumber \\
& ~~~
= - \alpha  b_k
 \nabla_\theta f(x; \theta_{k-1})^T \nabla_\theta f(z_k; \theta_{k-1}) 
+ O \left(  (\frac{\alpha}{n})^2  \right),
\label{eq:uk-uk-1-1}
\end{align}
where we treat $ L_f$ as $O(1)$ constant, and the same with other constants which depend on the infinity norm of derivatives of $f$. 

We analyze how $ \nabla_\theta f(x; \theta_{k-1})^T \nabla_\theta f(z_k; \theta_{k-1}) $ differs from
$ \nabla_\theta f(x; \theta_{0})^T \nabla_\theta f(z_k; \theta_{0}) $. For any $x \in \calX$,
\[
\nabla_\theta f(x; \theta_{k-1})
= \nabla_\theta f(x; \theta_{0})+ O( \| \theta_{k-1} - \theta_0\|),
\]
and by \eqref{eq:|thetak-theta0|},
\[
\nabla_\theta f(x; \theta_{k-1})
= \nabla_\theta f(x; \theta_{0})+ O \left(  \frac{k-1}{n}\alpha \right).
\]
Thus, 
\begin{align*}
 \nabla_\theta f(x; \theta_{k-1})^T \nabla_\theta f(z_k; \theta_{k-1})
& = \left \langle  \nabla_\theta f(x; \theta_{0})+ O \left(  \frac{k-1}{n}\alpha \right),  \nabla_\theta f(x; \theta_{0})+ O \left(  \frac{k-1}{n}\alpha \right) \right\rangle  \\
& =  \nabla_\theta f(x; \theta_{0})^T \nabla_\theta f(z_k; \theta_{0}) +  O \left(  \frac{k-1}{n}\alpha \right).
 \end{align*}
 Back to \eqref{eq:uk-uk-1-1}, we have
\begin{align*}
 u_{k}(x) - u_{k-1}(x)  
 & = 
 - \alpha  b_k
\left( \nabla_\theta f(x; \theta_{0})^T \nabla_\theta f(z_k; \theta_{0}) +  O \left(  \frac{k-1}{n}\alpha \right) \right)
+ O \left(  (\frac{\alpha}{n})^2  \right) \\
& =
- \alpha  b_k
\nabla_\theta f(x; \theta_{0})^T \nabla_\theta f(z_k; \theta_{0}) +  O \left(  \frac{k-1}{n^2}\alpha^2  \right) 
+ O \left(  (\frac{\alpha}{n})^2  \right). 
\end{align*}
This give that 
\begin{align*}
u_n(x) - u_0(x) 
& =
- \alpha \sum_{k=1}^n   b_k
\nabla_\theta f(x; \theta_{0})^T \nabla_\theta f(z_k; \theta_{0})  
+ \sum_{k=1}^n O\left(  k \frac{\alpha^2}{n^2}  \right)  \\
& = 
 \alpha \int_\calX K_0(x,x') (\hat{p} - \hat{q})(x') dx' 
+  O ( \alpha^2),
\end{align*}
where recall that $K_0(x,x') = \nabla_\theta f(x; \theta_{0})^T \nabla_\theta f(x'; \theta_{0})  $
is the NTK at time zero. 
This proves that
\[
\hat{g}(x):=  \frac{1}{\alpha} ( u_n(x) - u_0(x)  )
=\int_\calX K_0(x,x') (\hat{p} - \hat{q})(x') dx' 
+  O ( \alpha).
\]
Comparing to the continuous time training dynamic, 
we see that $\alpha$ corresponds to training time $t$,
and with batch size 1 the SGD training 
the NTK approximation has the same $O(\alpha)$ error as with the continuous time GD training.

\section{Proofs and additional theoretical results in Section \ref{sec:theory}}\label{appsec:proof-sec-theory}

\subsection{Proofs in Subsection \ref{subsec:hatT}}

The proof  of Theorem \ref{thm:power}
uses the U-statistic concentration analysis, which was used in Theorem 3.5 in \cite{ChengXie2021}.
The analysis in \cite{ChengXie2021} is for the local RBF kernel,
and we need to extend to the general PSD kernel here.

The concentration argument is by Proposition \ref{prop:conc-TNT}.
Note that the concentration can be derived using the
boundedness \eqref{eq:Kernel-uniform-bound} alone,
while the Bernstein-type control here is sharper when the squared integrals upper bound $\nu$ is much smaller than 1.

\begin{proposition}[Concentration of $\hat{T}_{\rm NTK}$]\label{prop:conc-TNT}
Assuming \eqref{eq:Kernel-uniform-bound}, \eqref{eq:def-cn-balance} and the conditions (i) and (ii) in Theorem \ref{thm:power}, 

(1) Under $H_0$, 
when $0 < \lambda < 3\sqrt{ c \nu n }$,
w.p. $\ge 1- 3 e^{-\lambda^2/8}$,
$\hat{T} \le \frac{4}{ cn} + 4 \lambda \sqrt{ \frac{\nu}{cn} }$.

(2) Under $H_1$, 
when $0 < \lambda < 3\sqrt{ c \nu n }$,
w.p. $\ge 1- 3 e^{-\lambda^2/8}$,
$\hat{T}
 \ge   \delta_K  - 4 \lambda \sqrt{ \frac{\nu}{cn} }$.

\end{proposition}

\rev{The proof of Theorem \ref{thm:power} is a direct application of the proposition.}
\begin{proof}[Proof of Theorem \ref{thm:power}]
Note that condition \eqref{eq:cond-n} ensures that
\begin{equation}\label{eq:cond-n-2}
\max\{ \lambda_1, \lambda_2 \} < 3 \sqrt{\nu c n},
\quad
\frac{4}{cn} < 0.5 \delta_K, 
\quad
4 (\lambda_1 + \lambda_2) \sqrt{ \frac{\nu}{ cn }} <  0.5 \delta_K, 
\end{equation}
and the bounds in Proposition \ref{prop:conc-TNT} parts (1) and (2) hold with $\lambda_1$ and $\lambda_2$ respectively. 

To verify that $\Pr [\hat{T} > t_{\thres}] \le \alpha_{\rm level}$ under $H_0$:
Observe that $3 e^{-\lambda_1^2/8} = \alpha_{\rm level}$ by the definition of $\lambda_1$,
and then the claim follows  by Proposition \ref{prop:conc-TNT} Part (1) since $\lambda_1 < 3 \sqrt{\nu c n}$.

To bound $\Pr [\hat{T} \le t_{\thres}] $ under $H_1$:
Since $\lambda_2 < 3 \sqrt{\nu c n}$, 
by Proposition \ref{prop:conc-TNT} Part (1),
the claim holds if 
\begin{equation}
t_{\thres} =   \frac{4}{cn } + 4 \lambda_1 \sqrt{ \frac{\nu}{cn} } < \delta_K - 4 \lambda_2 \sqrt{ \frac{\nu}{cn }}, 
\end{equation}
which is guaranteed by \eqref{eq:cond-n-2}. 
\end{proof}

\begin{remark}[Asymptotic choice of $t_{\thres}$]\label{rk:asymptotic-t-thres}
The optimal $t_{\thres}$ in Theorem \ref{thm:power} as the $(1-\alpha_{\rm level})$-quantile of the distribution of $\hat{T}$ under $H_0$
can be obtained potentially analytically according to the limiting distribution of the MMD statistic:
The asymptotic distribution of (squared) empirical MMD statistic has been derived using the spectral decomposition of the (centered) kernel function 
$\tilde{k}(x,x') := K(x,x') -  \E_{y \sim p} K(x,y) - \E_{y \sim p} K(y,x') + \E_{y,y' \sim p} K(y,y')$ in  \cite{gretton2012kernel, cheng2020two}, among others,
 following techniques in  Chapter 6 in \cite{serfling1981approximation}.
Specifically, by Theorem  3.3  in \cite{cheng2020two}, 
as $n=n_X + n_Y \to \infty$ and $n_X / n \to \rho_X \in (0,1)$,
$n\hat{T}$ under $H_0$, $q=p$, converges in distribution to the weighted $\chi^2$ distribution 
$\sum_{k=1}^\infty  \tilde{\lambda}_{k} \xi_k^2$,  
where
$\xi_k \sim {\cal N} \left(0,1/\rho_X + 1/(1-\rho_X) \right)$ i.i.d, 
and $  \tilde{\lambda}_{k} \ge 0$ are the eigenvalues of  the integral operator with kernel $\tilde{k}(x,x')$ in $L^2(\calX, p(x) dx)$.
This provides the asymptotic value of the quantile of $\hat{T}$ under $H_0$,
when the eigenvalues are computable, which can be useful, e.g., for low-dimensional data.  
\end{remark}

\begin{proof}[Proof of Proposition \ref{prop:conc-TNT}]
The proof follows the approach in Proposition. 3.4 in \cite{ChengXie2021}.
By definition, 

\begin{equation}
\label{eq:hatT-2}
\begin{split}
\hat{T}
& := 
\frac{1}{n_X^2}
\sum_{i,j =1}^{n_X}
K(x_i, x_j )
+ 
\frac{1}{n_Y^2 }
\sum_{i,j=1}^{n_Y}
K(y_i , y_j )
- \frac{2}{n_X n_Y}
\sum_{i=1}^{n_X}
\sum_{j=1}^{n_Y}
K(x_i , y_j),
 \end{split}
\end{equation}
and equivalently,
\begin{align}
\hat{T} & = \hat{T}_{X,X} + \hat{T}_{Y,Y} -2 \hat{T}_{X,Y},   \\
 \hat{T}_{X,X} & = \frac{1}{n_X^2}
\sum_{i,j =1}^{n_X}
K(x_i, x_j ),
\quad 
 \hat{T}_{Y,Y} = \frac{1}{n_Y^2}
\sum_{i,j =1}^{n_Y}
K( y_i, y_j ),
\quad 
 \hat{T}_{X,Y} =
\frac{1}{n_X n_Y}
\sum_{i=1}^{n_X}
\sum_{j=1}^{n_Y}
K(x_i , y_j).
\end{align}
The terms $\hat{T}_{X,X}$ and $\hat{T}_{Y,Y}$  contain diagonal entries of the kernel matrix which have different marginal distributions from the off-diagonal entries. 
Define
\[
D_X: = \frac{1}{n_X} \sum_{i=1}^{n_X} K(x_i,x_i),
\quad
V_{X,X} :=  \frac{1}{n_X(n_X-1)}\sum_{i \neq j, i,j=1}^{n_X} K(x_i,x_j),
\]
then
\begin{equation}\label{eq:hatTXX-2}
\hat{T}_{X, X} 
= \frac{1}{n_X} D_X + (1-\frac{1}{n_X}) V_{X,X}
= V_{X,X} + \frac{1}{n_X}( D_X - V_{X,X}).
\end{equation}
Observe that
\begin{align*}
|V_{X,X}| 
& \le \frac{1}{n_X(n_X-1)}\sum_{i \neq j, i,j=1}^{n_X} |K(x_i,x_j) |  \\
& \le \frac{1}{n_X(n_X-1)}\sum_{i \neq j, i,j=1}^{n_X} \sqrt{K(x_i,x_i) }\sqrt{ K(x_j,x_j) }  
\quad \text{($K(x,x')$ is PSD)}\\
& \le  \frac{1}{n_X(n_X-1)}\sum_{i \neq j, i,j=1}^{n_X} \frac{1}{2} ( K(x_i,x_i)  + K(x_j, x_j) ) \\
& = \frac{1}{n_X }\sum_{ i=1}^{n_X}   K(x_i,x_i)   = D_X,
\end{align*}
and, in addition, by \eqref{eq:Kernel-uniform-bound},
\[
0 \le D_X \le 1.
\]
Thus \eqref{eq:hatTXX-2} gives that 
\begin{equation}\label{eq:hatTXX-3}
V_{X,X} \le \hat{T}_{X, X}  \le V_{X,X} + \frac{2}{n_X} D_X \le V_{X,X} + \frac{2}{n_X} .
\end{equation}
The random variable $V_{X,X}$ is a U-statistic, 
where for $i \neq j$,
\[
\E  K(x_i,x_j) = \E_{x \sim p, y \sim p} K(x,y),
\]
and by condition (ii), 
\[
\text{Var} (K(x_i,x_j)) \le 
\E_{x \sim p, y\sim p} K(x,y)^2  = \nu_{p,p} \le \nu.
\]
As for the boundedness of the r.v. $K(x_i,x_j)$, by \eqref{eq:Kernel-uniform-bound}, 
\[
|K(x_i,x_j) | \le 1 = L.
\]
By the de-coupling of U-statistic in Proposition. 3.4 in \cite{ChengXie2021},
we obtain the Bernstein-type control of the tail probability, 
that is
\[
\Pr \left[ V_{X,X} - \E_{x \sim p, y \sim p} K(x,y) > t \right ]
\le  \exp\{ - \frac{ \frac{n_X-1}{2} t^2}{2 \nu + \frac{2}{3} t L} \},
\quad \forall t >0.
\]
Let $t = \lambda  \sqrt{ \frac{\nu}{ n_X -1} }$,
to obtain the sub-Gaussian tail we need $t L < 3 \nu$, that is, $t < 3 \nu$ by that $L=1$.
This gives that when $0 < \lambda  < 3 \sqrt{\nu (n_X-1)}$,
\[
\Pr \left[ V_{X,X} - \E_{x \sim p, y \sim p} K(x,y)  > \lambda \sqrt{ \frac{\nu }{ n_X -1} } \right]
\le \exp\{ - \frac{ (n_X-1) t^2}{8 \nu}  \}
= e^{-\lambda^2/8}.
\]
The same holds for
$\Pr \left[ V_{X,X} - \E_{x \sim p, y \sim p} K(x,y)  < - \lambda \sqrt{ \frac{\nu }{ n_X-1} } \right]$. 
Meanwhile, by \eqref{eq:def-cn-balance},
\[
c n \le n_X - 1.
\]
Together with \eqref{eq:hatTXX-3},
this gives that when $0 < \lambda  < 3 \sqrt{\nu cn }$,
\begin{equation}\label{eq:conc-TXX}
\begin{split}
\hat{T}_{X,X} 
& \le   \E_{x \sim p, y \sim p} K(x,y)   + \lambda \sqrt{ \frac{\nu }{ cn} } + \frac{2}{ cn } ,
\quad \text{w.p. $ \ge 1- e^{-\lambda^2/8}$}, \\
\hat{T}_{X,X} 
& \ge  \E_{x \sim p, y \sim p} K(x,y)   - \lambda \sqrt{ \frac{\nu }{ cn} },
\quad \text{w.p. $ \ge 1- e^{-\lambda^2/8}$}.
\end{split}
\end{equation}

The similar bound can be proved for $\hat{T}_{Y,Y}$, 
by defining $D_Y$ and $V_{Y,Y}$ similarly,
and  using that  $\nu_{q,q} \le \nu$
and $ cn \le n_Y- 1$.

To analyze the concentration of $\hat{T}_{X,Y}$,
which consists of the summation over the $n_X$-by-$n_Y$ array,
the de-coupling argument gives that 
for $M := \min\{n_X, n_Y\}$, and any $0 < \lambda < 3 \sqrt{ \nu M}$, 
\[
\Pr \left[ 
\hat{T}_{X,Y}  >   \E_{x \sim p, y \sim q} K(x,y)   + \lambda \sqrt{ \frac{\nu }{ M} }  
\right]
\le  e^{-\lambda^2/8}, 
\]
and same for 
$\Pr \left[ 
\hat{T}_{X,Y}  <   \E_{x \sim p, y \sim q} K(x,y)   - \lambda \sqrt{ \frac{\nu }{ M} }  
\right]$.
By that $cn \le M$,
when $0 < \lambda  < 3 \sqrt{\nu cn }$,
\begin{equation}\label{eq:conc-TXY}
\begin{split}
\hat{T}_{X,Y} 
& \le   \E_{x \sim p, y \sim q} K(x,y)   + \lambda \sqrt{ \frac{\nu }{ cn} },
\quad \text{w.p. $ \ge 1- e^{-\lambda^2/8}$}, \\
\hat{T}_{X,Y} 
& \ge  \E_{x \sim p, y \sim q} K(x,y)   - \lambda \sqrt{ \frac{\nu }{ cn} },
\quad \text{w.p. $ \ge 1- e^{-\lambda^2/8}$}.
\end{split}
\end{equation}

Finally, to prove Part (1) of the proposition,
use the upper bound in \eqref{eq:conc-TXX}, 
the corresponding upper bound for $\hat{T}_{YY}$,
and the lower bound in \eqref{eq:conc-TXY}. 
This gives that,
when $0 < \lambda  < 3 \sqrt{\nu cn }$,
 under the intersection of the three good events,
 which happens w.p. $\ge 1- 3 e^{-\lambda^2/8}$,
 we have that 
 \[
 \hat{T}_{X,X}
 +  \hat{T}_{Y,Y} 
 -2 \hat{T}_{X,Y}
 \le 
( \E_{x \sim p, y \sim p} +  \E_{x \sim q, y \sim q} -   2   \E_{x \sim p, y \sim q} )K(x,y)   
    +  4 \lambda \sqrt{ \frac{\nu }{ cn} } + \frac{4}{ cn },
 \]
where the first term vanishes since $p=q$ under $H_0$.
To prove part (2),
use the lower bounds in \eqref{eq:conc-TXX}, 
in the counterpart of \eqref{eq:conc-TXX} for $\hat{T}_{YY}$,
and the upper bound in \eqref{eq:conc-TXY}.
\end{proof}

\subsection{Proof of Theorem \ref{thm:power-asym-testonly-boot}}\label{appsubsec:proof-sec-3.2}

In the proof of Theorem \ref{thm:power-asym-testonly-boot} and \ref{thm:power-asym} which involves training and testing splitting,
we use subscript $_{(1)}$ to denote the randomness over $\calD_{tr}$,
and subscript $_{(2)}$  that over $\calD_{te}$, possibly conditioned on $\calD_{tr}$.
We use the notations $\Pr_{(i)}$, $\E_{(i)}$ and ${\rm Var}_{(i)}$, for $i=1, 2$. 
We say $E$ is a good event in $\Pr_{(1)}$ which happens w.p. $\ge 1- \delta$ in $\Pr_{(1)}$
 if $\Pr_{(1)}[ E^c] \le \delta$, where $0< \delta < 1$ is a small number.

 Theorem \ref{thm:power-asym-testonly-boot} is based on Lemma \ref{lemma:conc-hatC} which establishes the concentration of  the conditional expectation $\E [\hat{T}_a | \calD_{tr} ]$, and Proposition \ref{prop:conc-hatT-a-(2)} on the concentration of $\hat{T}_a$ under  good events of $\calD_{tr}$.

\begin{proof}[Proof of Theorem \ref{thm:power-asym-testonly-boot}]

We first consider under $H_0$, where $\delta_K = 0$.
Let $\gamma = 8 \delta$, and applying Lemma \ref{lemma:conc-hatC} with $\lambda_{(1)}$ such that 
\[
e^{-\lambda_{(1)}^2/4} = \delta,
\]
which gives the same value of $\lambda_{(1)}$ as in the statement of the theorem.
We have that there is a good event $E_{1}$ in $\Pr_{(1) }$, which happens w.p. $\ge 1 - 4\delta$,
such that under $E_1$, 
\begin{equation}\label{eq:E1-bound}
\hat{C} \le \delta_K + 4 \lambda_{(1)} \sqrt{\frac{\nu}{ c_a n}} =  4 \lambda_{(1)} \sqrt{\frac{\nu}{ c_a n}},
\end{equation}
and this requires 
\begin{equation}\label{eq:cond-pf-1}
\lambda_{(1)} < 3 \sqrt{\nu c_a n}.
\end{equation}
Applying Proposition \ref{prop:conc-hatT-a-(2)} (1), 
there is another good event $E_{2}$ in $\Pr_{(1) }$, which happens w.p. $\ge 1 - 4\delta$,
such that under $E_2$, 
\begin{equation}\label{eq:E2-bound}
\Pr_{(2)} \left[  \hat{T}_a > \hat{C}  + 4 \lambda_{(2),1} \sqrt{ \frac{1.1\nu}{c_a n}} \right] 
\le 4 e^{-\lambda_{(2),1}^2/4},
\end{equation}
as long as
\begin{equation}\label{eq:cond-pf-2}
\lambda_{(2),1}  < 3 \sqrt{ 1.1 \nu c_a n}, 
\quad \sqrt{ \log(1/\delta)/(2 c_a n) } =\sqrt{ \lambda_{(1)}^2/(8 c_a n) } 
\le 0.1 \nu.
\end{equation}
We thus set 
\[
 4 e^{-\lambda_{(2),1}^2/4} = \alpha_{\level},
\quad
t_{\thres} = 4 \lambda_{(1)} \sqrt{\frac{\nu}{ c_a n}} + 4 \lambda_{(2),1} \sqrt{ \frac{1.1\nu}{c_a n}},
\]
which gives the same values of $\lambda_{(2),1}$
and $t_{\thres}$ as in the statement of the theorem.
Then, under the intersection event $E_1 \cap E_2$ which happens w.p. $\ge 1-8\delta = 1-\gamma$ in $\Pr_{(1)}$, combining \eqref{eq:E1-bound} and \eqref{eq:E2-bound} gives that
\[
\Pr [  \hat{T}_a  > t_{\thres} ] \le \alpha_{\level}.
\]

Next, under $H_1$, similarly, there are good events $E_1'$ and $E_2'$,
the intersection of which  happens w.p. $\ge 1-\gamma$ in $\Pr_{(1)}$,
and under $E_1' \cap E_2'$,
\[
\hat{C} \ge \delta_K - 4 \lambda_{(1)} \sqrt{\frac{\nu}{ c_a n}},
\]
and
\[
\Pr_{(2)} \left[ \hat{T}_a < \hat{C}  - 4 \lambda_{(2),2} \sqrt{ \frac{1.1\nu}{c_a n}} \right] 
\le 4 e^{-\lambda_{(2),2}^2/4},
\]
and this requires 
\begin{equation}\label{eq:cond-pf-3}
\lambda_{2,(2)}  < 3 \sqrt{ 1.1 \nu c_a n}.
\end{equation}
This means that the Type-II error bound under $H_1$ in the theorem holds as long  as 
\begin{equation}\label{eq:cond-pf-4}
 \delta_K - 4 \lambda_{(1)} \sqrt{\frac{\nu}{ c_a n}} - 4 \lambda_{2,(2)} \sqrt{ \frac{1.1\nu}{c_a n}}
  > t_{\thres} .
\end{equation}
Collecting the needed requirements \eqref{eq:cond-pf-1} \eqref{eq:cond-pf-2} \eqref{eq:cond-pf-3} \eqref{eq:cond-pf-4}, 
and they are satisfied by \eqref{eq:cond-n-testonly-boot} and the assumption of the theorem. 
\end{proof}

In both Lemma \ref{lemma:conc-hatC} and Proposition \ref{prop:conc-hatT-a-(2)}, suppose that
\eqref{eq:Kernel-uniform-bound}, 
\eqref{eq:def-cn-balance-split} and the conditions (i) and (ii) in Theorem \ref{thm:power} hold.
We define  the witness function of exact NTK MMD as
\begin{equation}\label{eq:def-gNTK}
\hat{g}_{\NTK} (x) := \int_\calX K(x,x')( \hat{p}_{(1)}- \hat{q}_{(1)})(x') dx'.
\end{equation}

\begin{lemma}\label{lemma:conc-hatC}
Denote the conditional expectation $\E [\hat{T}_a | \calD_{tr} ]$ as 
\begin{equation}\label{eq:def-hatC}
\hat{C} := \int_{\calX} \hat{g}_{\NTK} (x) (p-q)(x) dx,
\end{equation}
then for any $ 0 < \lambda_{(1)}  < 3 \sqrt{\nu c_a n}$,
\[
\Pr_{(1)} \left[ \hat{C} - \delta_K > 4 \lambda_{(1)} \sqrt{ \frac{\nu}{ c_a n} } \right] \le 4 e^{-\lambda_{(1)}^2/4},
\]
and same with $\Pr_{(1)} [ \hat{C} - \delta_K < -4 \lambda_{(1)} \sqrt{ \frac{\nu}{ c_a n} } ] $.
\end{lemma}

\begin{proposition}\label{prop:conc-hatT-a-(2)}
Suppose $ 0< \delta < 1$ and $ \sqrt{ \log(1/\delta)/(2 c_a n) }\le 0.1 \nu$, then
under both $H_0$ and $H_1$, 
 there is a good event which happens w.p. $\ge 1- 4\delta$ over the randomness of $\calD_{tr}$, under which,  conditioning on $\calD_{tr}$,

(1) Under $H_0$,
$\Pr_{(2)} [ \hat{T}_a > \hat{C}  + 4 \lambda \sqrt{ \frac{1.1\nu}{c_a n}} ] \le 4 e^{-\lambda^2/4}$ 
if $0 < \lambda  < 3 \sqrt{ 1.1 \nu c_a n}$;

(2) Under $H_1$, 
$\Pr_{(2)} [ \hat{T}_a < \hat{C}  - 4 \lambda \sqrt{ \frac{1.1\nu}{c_a n}} ] \le 4 e^{-\lambda^2/4}$ 
if $0 < \lambda  < 3 \sqrt{ 1.1 \nu c_a n}$.
\end{proposition}
\begin{proof}[Proof of Proposition \ref{prop:conc-hatT-a-(2)}]
In this proof we write $\hat{g}_{\NTK}$ defined in \eqref{eq:def-gNTK} as $\hat{g}$ for shorthand notation. 
We have that $\hat{g} = \hat{g}_X - \hat{g}_Y$, 
where
\begin{equation}
\begin{split}
\hat{g}_X(x) := \int_\calX K(x,x') \hat{p}_{(1)} (x') dx' 
&= \frac{1}{n_{X,(1)}} \sum_{i=1}^{n_{X,(1)}} K(x, x_{i}^{(1)}), \\
\hat{g}_Y(x) := \int_\calX K(x,x') \hat{q}_{(1)} (x') dx'
&= \frac{1}{n_{Y,(1)}} \sum_{i =1}^{n_{Y,(1)}} K(x, y_{i}^{(1)}),
\end{split}
\end{equation}
and both $\hat{g}_X$ and $\hat{g}_Y$ are determined by $\calD_{tr}$. 
By definition, 
\begin{align}
\hat{T}_a 
& = \int_\calX ( \hat{g}_X  -  \hat{g}_Y )(x) (\hat{p}_{(2)} - \hat{q}_{(2)} ) (x) dx  \nonumber \\
& = 
\frac{1}{n_{X,(2)}} \sum_i \hat{g}_X( x_{i}^{(2)})
- \frac{1}{n_{Y,(2)}} \sum_i \hat{g}_X( y_{i}^{(2)})
- \frac{1}{n_{X,(2)}} \sum_i \hat{g}_Y( x_{i}^{(2)})
+ \frac{1}{n_{Y,(2)}} \sum_i \hat{g}_Y( x_{i}^{(2)}) \nonumber \\
& := S_{X,X} - S_{X,Y} - S_{Y,X} + S_{Y,Y}. \label{eq:def-SXX-four}
\end{align}
Conditioning on a realization of $\calD_{tr}$,
due to the independence of $\calD_{te}$ from $\calD_{tr}$, 
the four terms in \eqref{eq:def-SXX-four} are independent sums of random variables over the randomness of $\calD_{te}$.
Again, we analyze the concentration of these four terms respectively, conditioned on $\calD_{tr}$ and we will restrict to good events in $\Pr_{(1)}$. 

We start from $S_{X,X}$.
Again by \eqref{eq:Kernel-uniform-bound},  we have $|\hat{g}_X(x)| \le 1$ for any $x \in \calX$. 
Meanwhile, $\forall x \in \calX$,
\[
\hat{g}_X( x)^2 = \left( \frac{1}{n_{X,(1)}} \sum_i K(x, x_{i}^{(1)}) \right)^2
\le \frac{1}{n_{X,(1)}} \sum_i K(x, x_{i}^{(1)})^2,
\]
and thus, conditioning on $\calD_{tr}$,
\begin{equation}\label{eq:def-hatmu-pX}
\text{Var}_{(2)} ( \hat{g}_X( x_{i}^{(2)}))
\le 
\E_{ x \sim p} \hat{g}_X(x)^2
\le 
 \frac{1}{n_{X,(1)}} \sum_i  \E_{ x \sim p}  K(x, x_{i}^{(1)})^2
 = \frac{1}{n_{X,(1)}} \sum_i   \psi_p(x_{i}^{(1)}) =: \hat{\nu}_{ p, X}, 
\end{equation}
where we define
\[
\psi_p( x'): = \int_\calX K( x, x')^2 p(x) dx,
\]
and $\hat{\nu}_{ p, X}$ is a random variable determined by $\calD_{tr}$.
One can verify that by restricting to large probability event in $\Pr_{(1)}$,
$\hat{\nu}_{ p, X}$ concentrates at the mean value 
\begin{equation}\label{eq:E1hatnu-bound}
\E_{(1)} \hat{\nu}_{ p, X} =  \int_\calX \psi_p(x') p(x') dx'
= \int_\calX \int_\calX K( x, x')^2 p(x) dx p(x') dx'  = \nu_{p,p} \le \nu.
\end{equation}
Specifically, \eqref{eq:Kernel-uniform-bound} implies that $0 \le \psi_p( x') \le 1$,
and then by Hoeffding's inequality, 
\[
\Pr_{(1) } [   \hat{\nu}_{ p, X} - \E_{(1)} \hat{\nu}_{ p, X}  >  t ]
\le e^{- 2 n_{X,(1)} t^2} 
\le e^{- 2 c_a n  t^2} ,
\quad \forall t >0.
\]
Let $e^{- 2 c_a n  t^2} = \delta$, where $\delta$ is as in the statement of the proposition, then w.p. $\ge 1- \delta$ in $\Pr_{(1)}$, 
\begin{equation}\label{eq:good-event-EXX-bound}
\hat{\nu}_{p,X} \le  \E_{(1)} \hat{\nu}_{ p, X}   + t =  \E_{(1)} \hat{\nu}_{ p, X}  + \sqrt{ \frac{\log(1/\delta) }{ 2c_a n}}
\le \nu + 0.1 \nu,
\end{equation}
and the last inequality is by \eqref{eq:E1hatnu-bound} and the condition of the proposition. 
We call this good event  $E_{X,X}$  in $\Pr_{(1)}$,
under which \eqref{eq:good-event-EXX-bound} holds. 

Back to $S_{X,X}$, we have that under $E_{X,X}$ in in $\Pr_{(1)}$,
 and conditioning on the realization of $\calD_{tr}$, 
 $\hat{g}_X( x_{i}^{(2)})$ as r.v. in $\Pr_{(2) }$  are bounded as $|\hat{g}_X( x_{i}^{(2)}) | \le 1$;
 Meanwhile, by \eqref{eq:def-hatmu-pX} and \eqref{eq:good-event-EXX-bound},
\[
\text{Var}_{(2)}( \hat{g}_X( x_{i}^{(2)}) )  \le \hat{\nu}_{p,X} \le 1.1\nu.
\]
Then the classical Bernstein gives that 
 $\forall  0 < \lambda < 3 \sqrt{1.1 \nu n_{X,(2)}}$,
\[
\Pr_{(2)} \left[ S_{X,X} - \E_{(2)} S_{X,X} > \lambda \sqrt{\frac{ 1.1 \nu}{ n_{X,(2)} }} \right], \,
\Pr_{(2)} \left[ S_{X,X} - \E_{(2)} S_{X,X} < - \lambda \sqrt{\frac{1.1 \nu}{ n_{X,(2)} }} \right] 
\le e^{-\lambda^2/4}.
\]
By that $n_{X,(2)} \ge c_a n $, we have that $\forall  0 < \lambda < 3 \sqrt{1.1 \nu c_a n }$,
under the good event $E_{X,X}$ which happens w.p. $\ge 1- \delta$ in $\Pr_{(1)}$ and conditioning on $\calD_{tr}$,
\begin{equation}\label{eq:Pr2-SXX-bound}
\Pr_{(2)} \left[ S_{X,X} - \E_{(2)} S_{X,X} > \lambda \sqrt{\frac{ 1.1 \nu}{ c_a n  }} \right], \,
\Pr_{(2)} \left[ S_{X,X} - \E_{(2)} S_{X,X} < - \lambda \sqrt{\frac{1.1 \nu}{ c_a n }} \right] 
\le e^{-\lambda^2/4}.
\end{equation}

Similarly, we can show that, 
there are good events $E_{X,Y}$, $E_{Y,X}$, and $E_{Y,Y}$ over randomness of $\calD_{tr}$,
where each happens in $\Pr_{(1)}$ w.p. $\ge 1- \delta$, 
and under which 
the similar bound as \eqref{eq:Pr2-SXX-bound} holds   for $S_{X,Y}$, $S_{Y,X}$, and $S_{Y,Y}$ respectively
as long as  $0 < \lambda < 3 \sqrt{1.1 \nu c_a n }$.
Thus,
 under the intersection of the four good events, which happens in $\Pr_{(1)}$ w.p. $\ge 1- 4\delta$, 
\[
\Pr_{(2)} \left[ \hat{T}_a - \E_{(2)} \hat{T_a} > 4 \lambda \sqrt{\frac{ 1.1 \nu}{ c_a n  }} \right], \,
\Pr_{(2)} \left[ \hat{T}_a - \E_{(2)} \hat{T_a} < - 4\lambda \sqrt{\frac{ 1.1 \nu}{ c_a n  }} \right]
\le 4 e^{-\lambda^2/4}.
\]
The above holds under both $H_0$ and $H_1$.
Finally, by that $\E_{(2)} \hat{T}_a = \hat{C}$ as defined in \eqref{eq:def-hatC},
 this proves parts (1) and (2) of the proposition.
\end{proof}
\begin{proof}[Proof of Lemma \ref{lemma:conc-hatC}]
Note that $\hat{C}$ is a random variable over the randomness of $\calD_{tr}$ only. 
By definition,
\[
\hat{C} = \int_\calX \int_\calX K(x,x')( \hat{p}_{(1)}- \hat{q}_{(1)})(x') dx' (p-q)(x) dx 
= \int_\calX (\varphi_p - \varphi_q)(x') ( \hat{p}_{(1)}- \hat{q}_{(1)})(x') dx',
\]
where 
\[
\varphi_p(x') := \int_\calX K(x,x') p(x) dx, 
\quad  
\varphi_q(x') := \int_\calX K(x,x') q(x) dx.
\]
Because only $n_{X,(1)}$ and $n_{Y,(1)}$ are involved here, in this proof we write $n_{X,(1)}$ as $n_X$ 
and  $n_{Y,(1)}$ as $n_Y$ for notation convenience,
and we also denote samples from $X_{(1)}$ and $Y_{(1)}$ by $x_i$ and $y_i$ respectively.
By \eqref{eq:def-cn-balance-split}, we then have
\begin{equation}\label{eq:pf-ca-n}
 n_X, \, n_Y  \ge c_a n.
\end{equation}

We then equivalently write $\hat{C}$ as
\begin{align}
\hat{C} 
& =
 \frac{1}{n_{X} }\sum_{i=1}^{ n_{X} }
\varphi_p( x_{i}) 
- 
 \frac{1}{n_{X}}\sum_{i=1}^{ n_{X} }
\varphi_q( x_{i}) 
- \frac{1}{n_{Y}}\sum_{i=1}^{ n_{Y} }
\varphi_p( y_{i}) 
+
 \frac{1}{n_{Y}}\sum_{i=1}^{ n_{Y} }
\varphi_q( y_{i})  \nonumber \\
&
: = C_{X,X} - C_{X,Y} - C_{Y,X} + C_{Y,Y},
\end{align}
and we use concentration argument on the four terms respectively. 

Due to \eqref{eq:Kernel-uniform-bound},
\[
|\varphi_p(x) | \le 1, \quad |\varphi_q(x)| \le 1,  \quad \forall x \in \calX.
\]
Starting from $C_{X,X}$ which is an independent sum of i.i.d. rv's,
where $ | \varphi_p(x_i) | \le 1 :=L$;
By that 
\[
\varphi_p(x)^2
 = \left( \int_\calX K(x,x') p(x') dx'  \right)^2
\le  
\left( \int_\calX K(x,x')^2 p(x') dx' \right)  \left( \int_\calX p(x') dx'  \right)
=\int_\calX K(x,x')^2 p(x') dx'
\]
we have
\[
\text{Var}_{(1)}( \varphi_p(x_i)  ) \le \E_{x \sim p} \varphi_p(x)^2
\le 
\int_\calX  \int_\calX K(x,x')^2 p(x') dx'   p(x) dx = \nu_{p,p} \le \nu,
\]
where the last inequality is by condition (ii) in Theorem \ref{thm:power}.
The classical Bernstein then gives that $\forall  0 < \lambda < 3 \sqrt{\nu n_X}$,
\[
\Pr_{(1)} \left[ C_{X,X} - \E_{(1)} C_{X,X} > \lambda \sqrt{\frac{\nu}{ n_X}} \right], \,
\Pr_{(1)} \left[ C_{X,X} - \E_{(1)} C_{X,X} < - \lambda \sqrt{\frac{\nu}{ n_X}} \right] 
\le e^{-\lambda^2/4}.
\]

The similar bounds can be derived for $C_{X,Y}$, 
and for $C_{Y,X}$ and $C_{Y,Y}$ where $n_X$ is replaced with $n_Y$. 
By \eqref{eq:pf-ca-n}, this gives that 
when $ 0 < \lambda < 3 \sqrt{\nu c_a n } \le 3 \sqrt{\nu n_X}$ and $3 \sqrt{\nu n_Y}$, 
\[
\Pr_{(1)} \left[ \hat{C} - \E_{(1)} \hat{C}  > 4 \lambda \sqrt{\frac{\nu}{ c_a n}} \right],
\,
\Pr_{(1)} \left[ \hat{C} - \E_{(1)} \hat{C}  <- 4 \lambda \sqrt{\frac{\nu}{ c_a n}} \right]
\le 4 e^{- \lambda^2/4}.
\]
Observing that
$\E_{(1)} \hat{C}  =  \delta_K$ which is defined in \eqref{eq:def-delta-K} finishes the proof.
\end{proof}

 \subsection{Test power of $\hat{T}_a$ with full-bootstrap}
 \label{appsec:hatT-a-thres-theory}
 
 We derive here the testing power of the statistic $\hat{T}_a$ computed on split training/testing sets in Subsection \ref{subsec:hatT-a},
 with a theoretical choice of $t_{\thres}$, similar to as in Theorem \ref{thm:power}. 
 In practice, the full-bootstrap estimation of $t_{\thres}$ can obtain better power than the theoretical one.

%
%
%
%
%


\begin{proof}[Proof of Theorem \ref{thm:power-asym}]
Similar to the proof of Theorem \ref{thm:power}
by applying Proposition \ref{prop:conc-TNT-asym}. 
Due to that the upper bound of $\hat{T}_a$ under $H_0$ does not have the $\frac{4}{cn}$  term, 
c.f.  Proposition \ref{prop:conc-TNT-asym} Part (1)  (because the asymmetric kernel MMD is computed from an off-diagonal block
of the kernel matrix and the summation in $\hat{T}_a$ does not involve diagonal terms),
the value of $t_{\thres}$ does not have the $\frac{4}{cn}$ term,
and the condition \eqref{eq:cond-n-split} has one term less on the r.h.s. than  \eqref{eq:cond-n}. 
\end{proof}

\begin{proposition}[Concentration of $\hat{T}_{a}$]\label{prop:conc-TNT-asym}
Assuming \eqref{eq:Kernel-uniform-bound}, 
\eqref{eq:def-cn-balance-split} and the conditions (i) and (ii) in Theorem \ref{thm:power}, 

(1) Under $H_0$, 
 when $0 < \lambda < 3\sqrt{ c_a \nu n }$,
w.p. $\ge 1- 4 e^{-\lambda^2/8}$,
$
\hat{T}_a \le  4 \lambda \sqrt{ \frac{\nu}{c_a n} }$.

(3) Under $H_1$, 
 when $0 < \lambda < 3\sqrt{ c_a \nu n }$,
w.p. $\ge 1- 4e^{-\lambda^2/8}$,
$
\hat{T}_a  \ge  
\delta_K - 4 \lambda \sqrt{ \frac{\nu }{c_a n} }$.

\end{proposition}

The proof makes use of the independence of the four datasets 
$ X_{(1)}$, $X_{(2)}$, 
$ Y_{(1)}$ and  $Y_{(2)}$,
and the concentration of the double summation over the four blocks of the asymmetric  kernel matrix.

\begin{proof}[Proof of Proposition \ref{prop:conc-TNT-asym}]
By definition, 
\begin{equation}
\label{eq:hatT-2-asym}
\begin{split}
\hat{T}_{a}
& = 
\frac{1}{n_{X,(2)} n_{X,(1)}}
\sum_{i=1}^{n_{X,(2)}}
\sum_{j=1}^{n_{X,(1)}}
K(x_i^{(2)},x_j^{(1)})
- \frac{1}{n_{X,(2)} n_{Y,(1)}}
\sum_{i=1}^{n_{X,(2)}}
\sum_{j=1}^{n_{Y,(1)}}
K(x_i^{(2)}, y_j^{(1)})
\\
&  ~~~
- \frac{1}{n_{Y,(2)} n_{X,(1)}}
\sum_{i=1}^{n_{X,(1)}} \sum_{j=1}^{n_{Y,(2)}}
K(x_i^{(1)}, y_j^{(2)} )
+ 
\frac{1}{n_{Y,(2)} n_{Y,(1)}}
\sum_{i=1}^{n_{Y,(2)}}
\sum_{j=1}^{n_{Y,(1)}}
K(y_i^{(2)},y_j^{(1)}).
 \end{split}
\end{equation}
Then, equivalently, 
\begin{align}
\hat{T}_{a} 
& =  T_{X,X} - T_{Y,X} - T_{X,Y}  + T_{Y,Y}
\\
T_{X,X} 
&:= 
\frac{1}{n_{X,(2)} n_{X,(1)}}
\sum_{i=1}^{n_{X,(2)}}
\sum_{j=1}^{n_{X,(1)}}
K(x_i^{(2)},x_j^{(1)}),
\quad
T_{Y,X}
: =  \frac{1}{n_{X,(2)} n_{Y,(1)}}
\sum_{i=1}^{n_{X,(2)}}
\sum_{j=1}^{n_{Y,(1)}}
K(x_i^{(2)}, y_j^{(1)})
\\
T_{X,Y}& :=
 \frac{1}{n_{Y,(2)} n_{X,(1)}}
\sum_{i=1}^{n_{X,(1)}} \sum_{j=1}^{n_{Y,(2)}}
K(x_i^{(1)}, y_j^{(2)} ),
\quad
T_{Y,Y}:=
\frac{1}{n_{Y,(2)} n_{Y,(1)}}
\sum_{i=1}^{n_{Y,(2)}}
\sum_{j=1}^{n_{Y,(1)}}
K(y_i^{(2)},y_j^{(1)}).
\end{align}

We analyze the concentration of the four terms respectively, 
all similarly to the analysis of the ``$\hat{T}_{X,Y}$" term in the proof of Proposition \ref{prop:conc-TNT},
Specifically, for $T_{X,X}$:
Define $M:= \min\{ n_{X,(1)},  n_{X,(2)}\}$,
and by \eqref{eq:def-cn-balance-split}, 
\[
M \ge c_a n. 
\]
 
By that $\nu_{pp} \le \nu$ and that the kernel is bounded in absolute value by 1,
we have that
$\forall 0 < \lambda < 3\sqrt{ \nu M  }$, 
\[
\Pr \left[ T_{X,X} - \E_{x \sim p, y \sim p} K(x,y)  > \lambda \sqrt{ \frac{\nu }{ M} } \right],
\Pr \left[ T_{X,X} - \E_{x \sim p, y \sim p} K(x,y)  < - \lambda \sqrt{ \frac{\nu }{ M} } \right]
\le e^{-\lambda^2/8},
\]
and $M$ can be replaced to be $c_a n$ where the claim remains to hold. 
Similar bounds hold for $T_{Y,X}$, $T_{X,Y}$, $T_{Y,Y}$,
since 
\[
\min\{ n_{X,(1)},  n_{X,(2)},  n_{Y,(1)}, n_{Y,(2)}\}  
 \ge cn. 
\]

Putting together, 
to prove (1) under $H_0$, use the concentration bounds for the 4 quantities and under the joint good events, 
plus that ${\rm MMD}_K^2(p,q) = 0$.
Part (2) under $H_1$ is proved similarly. 
\end{proof}

\section{Experimental details \rev{and additional results}}

\subsection{Gaussian mean and covariance shifts}\label{appsub:exp-gaussian-more}

The neural network has 2 fully-connected (fc) layers, i.e. 1 hidden layer, and has the following architecture: the input data dimension $d=100$, the hidden layer width $m=512$,

~~~~~
fc ($d$, $m$) - softplus - fc ($m$, 1) - loss as in \eqref{eq:def-Loss}
 
where $({\rm f}_{in}, {\rm f}_{out})$ stand for dimensionality of input and output features respectively.

The network  mapping $f(x; \theta)$ can be equivalently written as
\begin{equation}\label{eq:2lnn-f}
f(x; \theta) = \sum_{k=1}^m a_k \sigma( w_k^T x + b_k), \quad \theta = \{(w_k, b_k, a_k)\}_{k=1}^m.
\end{equation}
The neural network parameters are initialized such that $a_k \sim \calN(0,{1}/{m})$,
$w_k \sim \calN(0, I_d)$, and $b_k = 0$.
For simplicity, we leave the 2nd layer parameters $a_k$ fixed after initialization and only train the 1st layer parameters $w_k$ and $b_k$.

\begin{remark}[Effective learning rate]\label{rk:effective-lr}
The network is trained for 1 epoch (1 pass of the training set) and batch-size 1, using basic SGD.
In the notation of Remark \ref{rk:SGD}, the theoretical learning rate $\alpha = 0.1$.
Note that the definition of loss  \eqref{eq:def-Loss} contains normalization $1/n_X$ and $1/n_Y$,
and here $n_{X,(1)} = n_{Y,(1)} = 100$.
Comparing to training objective which is usually defined as the summation (with out normalizing by sample size),
the effective learning rate here (lr) is $\alpha/100 = 10^{-3}$.
Using smaller values of lr produces similar results,
but note that reducing lr to be too small may cause numerical issue,
 due to that the deep learning programs use single precision floating point arithmetic. 
 \end{remark}

The testing powers are approximately computed over  $n_{\rm run}$ random replicas. 
For Figure \ref{fig:exp-gaussian},
the most right plot is produced by  $n_{\rm run}=200$, and all  other plots by  $n_{\rm run} = 500$.
\rev{In the most right plot, $n_{X, (1)} = n_{Y, (1)} = 250$, and the effective lr is $0.1/250 = 4\times 10^{-4}$. }

\subsection{Experiments of varying neural network hyperparameters}\label{appsub:exp-additional}

\rev{
We conducte additional experiments to investigate the influence of neural network architecture and training hyperparameters. 
}

$\bullet$ Different activation functions, network depths and widths

Table \ref{tab:exp-more-relu} shows that increasing the network depth can improve testing power,
and  changing from softplus to relu obtains similar results.
We also find in experiments that relu can obtain more robustness of testing power performance with respect to different weight initialization schemes. 
We observe that the performance with wider networks is generally better, though no longer sensitive beyond a certain $m$. 
Theoretically, the convergence to infinite-width limiting NTK may lead to further analysis of the discriminative power of the kernel to distinguish $p$ and $q$,
see the comments in Subsection \ref{subsec:exp-C2ST}.

\begin{table}
\centering
\small
\begin{tabular}[t]{ l  |  c c c     }
\hline
Neural network configuration $\backslash$ width $m$     	&   	 256        &    	512      &          1024 \\
\hline 
2-layer softplus  	& 	 82.0	      &		81.6      &           82.0 	\\
2-layer relu            	&	 79.8	      &        84.4      &            82.8 	\\
3-layer relu 	      	&	 85.8	      & 	88.4      &           91.0 	\\
\hline
\end{tabular}
\caption{
\rev{
NTK-MMD with relu activation, different width $m$, and more layers 
(to compare to Figure \ref{fig:exp-gaussian}, which is computed with 2 fc-layers, softplus activation, width $m$=512). 
Numbers in the table are testing power (in \%).
}
}\label{tab:exp-more-relu}
\end{table}

\begin{table}
\centering
\small
\begin{tabular}[t]{ l  |  c     }
\hline
     SGD configuration                   &           Test power of NTK-MMD \\
\hline
Batch-size = 1,   epoch= 1 (10)   & 		84.2 (85.0) \\
Batch-size = 20, epoch= 1 (10)   & 		82.8 (81.6) \\
\hline
\end{tabular}
\caption{
\rev{
NTK-MMD trained with different numbers of epochs and batch sizes. 
The example of  gaussian covariance shift in Section \ref{subsec:exp-gaussian}, $\rho = 0.12$. Test power (in \%) with epoch=1 outside brackets, with epoch=10 in brackets.
}
}\label{tab:exp-more-sgd}
\end{table}

$\bullet$ General SGD with varying batch-size, epochs, and batch-size 

Theoretically, the analysis covers general SGD (more than one epoch and different batch size): 
The proof in Appendix \ref{subsec:sgd-bound} generalizes to such cases because the residual error of the Taylor expansion of the network mapping $f(x;\theta)$ still applies. 

Empirically, we verify that the testing power of NTK-MMD is not sensitive to batch size nor a few more epochs, as illustrated in Table \ref{tab:exp-more-sgd}. 
This agrees with the theory that $\hat{T}_{net}$ computed with different batch-size and small number of epochs all approximate the exact NTK-MMD at time zero. 
In other experiments in the paper, we focus on batch-size =1 to show that NTK-MMD allows extremely small batch-size. Note that the advantage of NTK-MMD is particularly pronounced in the one-pass training, i.e., we can only visit the data in one-pass, which commonly appears in the streaming data setting.

\subsection{Computation of the exact NTK MMD}\label{appsub:exp-exact-more}

\begin{figure}
\begin{center}
\includegraphics[trim =  0 0 0 0, clip, height=.225\linewidth]{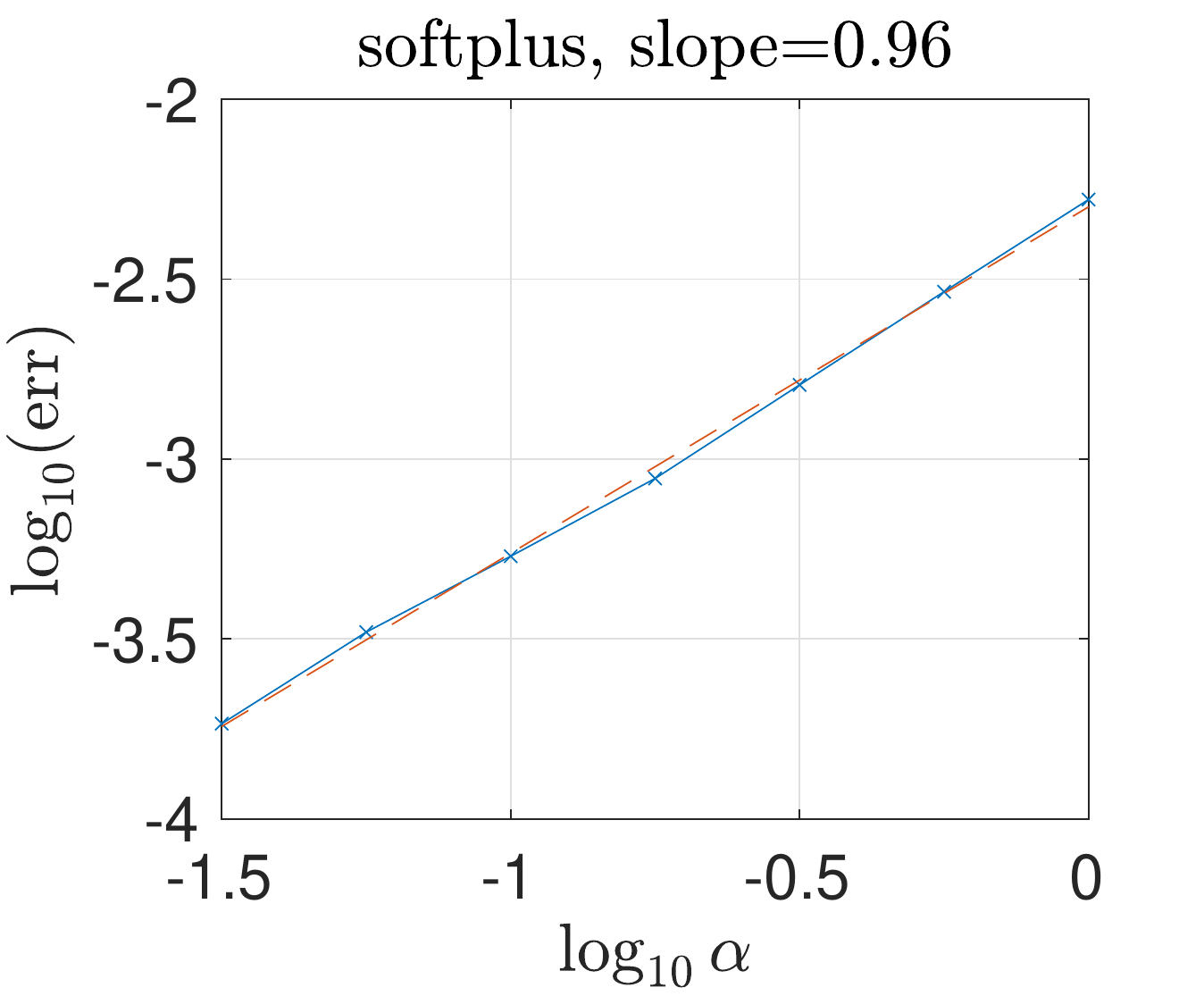} 
\includegraphics[trim =  0 0 0 0, clip, height=.225\linewidth]{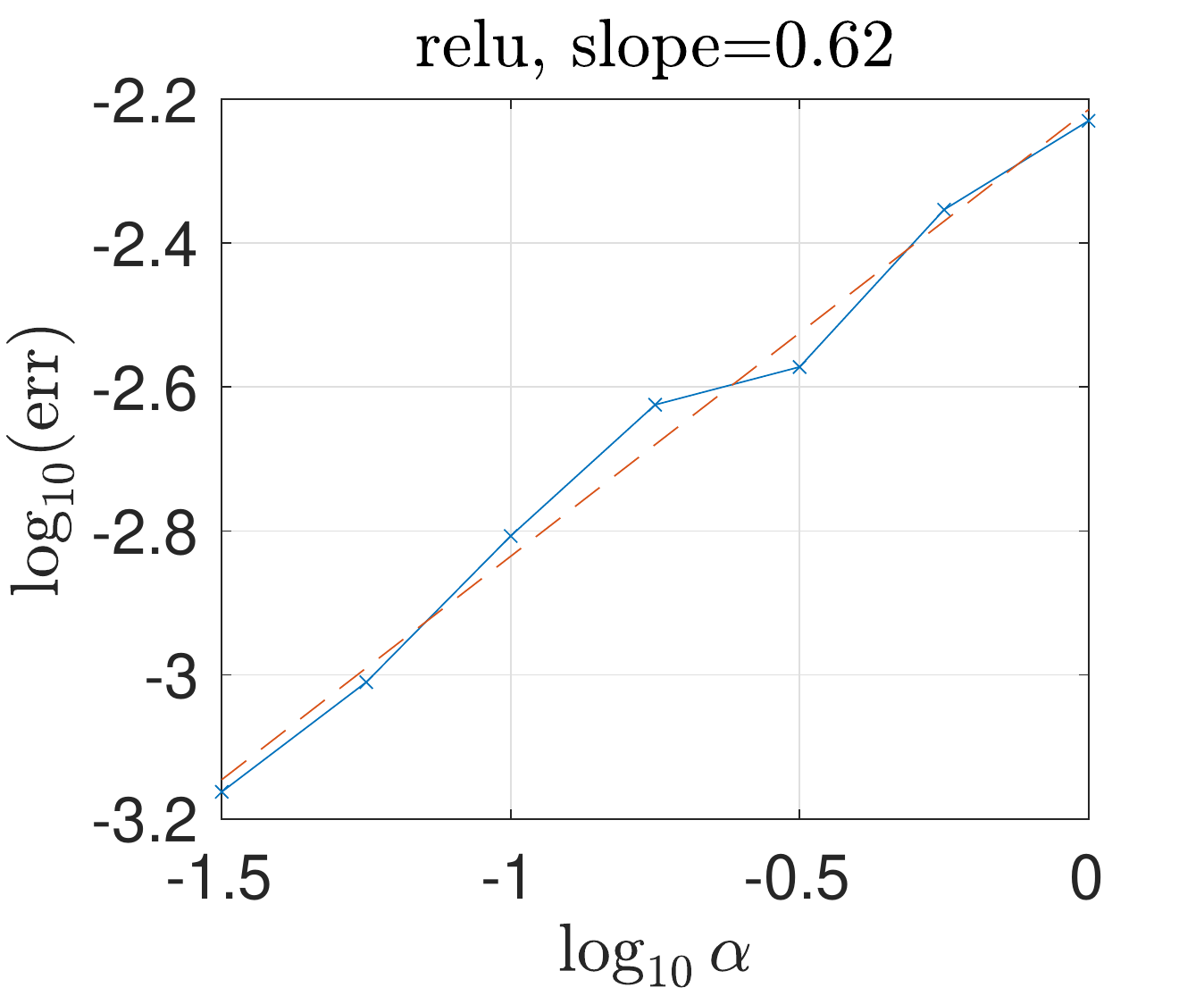}
\end{center}
\vspace{-10pt}
\caption{
\rev{
Numerical computation of the (relative) approximation error ${\rm err}=|\hat{T}_{\rm net}(t)-\hat{T}_{\rm NTK}|/|\hat{T}_{\rm NTK}|$.
 The network has 2-fc layers with softplus (or relu) activation and width $m=512$,
 and the simulated data distribution is Gaussian covariance shift in dimension 100. 
 }
}\label{tab:O(t)-error}
\end{figure}

The neural network setting is the same as in Subsection \ref{subsec:exp-gaussian}, 
and here we derive the expression of the NTK kernel at $t=0$, which was used to compute the  ``ntk1'' and ``ntk2'' statistics. 

For the network function as in \eqref{eq:2lnn-f},
\begin{equation}\label{eq:ntk-t0-expression}
K_0( x, x')
= \left( 
\sum_{k=1}^m a_k^2 \sigma'( w_k^T x + b_k )  \sigma'( w_k^T x' + b_k )
\right)
\left( 1+ x^T x'\right),
\end{equation}
where $\sigma(z) =\log (1 + e^z) $ is the softplus function, and is differentiable on $\R$. 
Thus the kernel for any pair of samples $x$ and $x'$ is analytically computable once the network parameters are initialized. 
In our experiments, we compute $K_0(x,x')$ as in \eqref{eq:ntk-t0-expression} with finite hidden-layer width $m$ and given realizations of the $t=0$ network parameters.

\subsection{Comparison to neural network classification tests}\label{appsub:exp-C2ST-more}

The network is fc 3-layer with relu activation and width $m=512$. Two C2ST baselines are trained with Adam and SGD respectively, and trained for 1 and 10 epochs. (By SGD, we mean vanilla SGD with constant step-size and no momentum.) NTK-MMD uses SGD, epoch $=1$. We also experiment under $H_0$ to verify that the Type-I error achieves $\alpha_{\rm level} =0.05$.

Since \cite{liu2020learning}  already compared C2ST’s with optimization-based linear time kernel tests, namely ME and SCF tests \cite{chwialkowski2015fast,jitkrittum2016interpretable}
 and showed that C2ST’s are generally better, we cite the results therein for comparison.

\subsection{MNIST distribution abundance change}\label{appsub:exp-mnist-more}

The neural network has two convolutional  (conv)  layers:

~~~~~
conv 5x5x1x16 - relu - maxpooling 2x2 

~~~~~
- conv 5x5x16x32 - relu - maxpooling 2x2  

~~~~~
- fc ( $\cdot$,128) - relu - fc (128, 1) - loss

where  the dimension of ${\rm f}_{in}$ in the 1st fc layer is by flattening the input feature, which gives ${\rm f}_{in}= 4^2\cdot 32$ in this case.

In the online training of the network, we use batch size = 1, theoretical lr $\alpha = 0.01$,
and SGD with momentum 0.9,
Adding momentum to SGD is common in neural network practice, 
and we adopt it here as to examine the behavior of the model: theoretically, under the NTK assumption,
we expect similar behavior with and without momentum in short-time training with SGD. 
As has been explained in Appendix \ref{appsub:exp-gaussian-more},
by that $n_{X,(1)} = n_{Y,(1)} = 10^3$, 
the effective lr is $\alpha/10^3= 10^{-5}$.

\subsection{Human activity change-point detection}\label{appsub:exp-change-point-more}

The (MSRC-12) Kinect gesture dataset consists of sequences of human skeletal body part movements (represented as body part locations) collected from 30 people performing 12 gestures. There are 18 sensors in total, and each sensor records the coordinates in the three-dimensional Cartesian coordinate system at each time.

The net MMD statistic is computed using a 2-layer fc network having 512 hidden nodes and softplus activation.
We use effective lr 0.0015 and SGD with momentum 0.9 in the one-pass training with batch-size one.

\subsection{Comparison to linear time MMD}\label{appsub:exp-linear-time-MMD}

The test power comparison of NTK-MMD, Gaussian kernel MMD and the linear-time version as in \cite[Section 6]{gretton2012kernel},
on the  example of MNIST data in Section \ref{subsec:exp-mnist} is given in Table \ref{tab:mnist-linear-mmd}.
On the examples in Section \ref{subsec:exp-gaussian} (Figure \ref{fig:exp-gaussian})
 linear-time gaussian MMD baseline gives inferior power (all less than 10\%, details omitted). 
This version of linear-time MMD only provides a global test statistic but not directly a witness function (to indicate where $p$ and $q$ differ), while NTK-MMD training obtains network witness function which approximates the kernel witness function of NTK.

As alternative linear-time MMD tests, 
the ME and SCF tests  \cite{chwialkowski2015fast,jitkrittum2016interpretable}
involve additional gradient-based optimization of model parameters 
and may not have optimization convergence guarantee for general data distributions.  
NTK-MMD has comparable computational and memory complexity to classification neural network tests (the order is the same, but only one epoch is needed and batch size can be as small as one), and has learning guarantee via NTK approximation as shown in Section \ref{subsec:ntk-approx-error} and Section \ref{sec:theory}.

\begin{table}
\centering
\small
\begin{tabular}[t]{ l | c c c c c c   }
\hline
Test statistics $\backslash$ $n_{tr}$          &  100       & 200         & 300   	  & 500 	 &    1000     &   2000	 \\
 \hline
{\it gmmd}  	         & 62.0      &	93.2    &    99.6     &      -       &       -        &        -	\\
{\it gmmd-lin }		&  7.0       & 	10.8     &   12.6     &     16.0   &      24.4   &      36.4 	\\
 {NTK-MMD}  	& 35.4     & 	 67.6     &   86.2    &       98.2  &     100.0 &       100.0	\\
 \hline
\end{tabular}
\caption{
\rev{
Testing power (in \%) of MNIST density departure example in Subsection \ref{subsec:exp-mnist}. 
{\it gmmd} is Gaussian kernel MMD, and {\it gmmd-lin} the linear-time version. 
Results of  {\it gmmd} for $n_{tr}$ greater than 300 are omitted due to slow computation.
}
}\label{tab:mnist-linear-mmd}
\end{table}

\end{document}